\pdfoutput=1

\documentclass[11pt]{article}

\usepackage[]{acl}

\usepackage{times}
\usepackage{latexsym}

\usepackage[T1]{fontenc}

\usepackage[utf8]{inputenc}

\usepackage{microtype}

\usepackage{inconsolata}

\usepackage{graphicx}
\usepackage{amssymb}
\usepackage{booktabs}
\usepackage{multirow}
\usepackage{inconsolata}
\usepackage{makecell}
\usepackage{float}
\usepackage{bm}
\usepackage{booktabs}
\usepackage{multirow}
\usepackage{colortbl}
\usepackage{xspace}
\usepackage{color}
\usepackage{paralist}
\usepackage{amsmath,amsfonts,amsthm}
\usepackage{CJK}
\usepackage{tcolorbox}
\usepackage{algorithm}
\usepackage{algorithmic}

\usepackage{subfigure}

\title{Attention-guided Self-reflection for Zero-shot Hallucination Detection in Large Language Models}

\author{\textbf{ Qiang Liu$^{1}$, Xinlong Chen$^{1}$, Yue Ding$^{1}$, Bowen Song$^{2}$, Weiqiang Wang$^{2}$, Shu Wu$^{1}$\thanks{Corresponding author}, Liang Wang$^{1}$} \\
$^1$New Laboratory of Pattern Recognition (NLPR), \\
  State Key Laboratory of Multimodal Artificial Intelligence Systems (MAIS),\\
  Institute of Automation, Chinese Academy of Sciences (CASIA) \\
$^2$Ant Group \\
  \texttt{\{qiang.liu, xinlong.chen, yue.ding\}@nlpr.ia.ac.cn}, \\
  \texttt{\{bowen.sbw,weiqiang.wwq\}@antgroup.com}, \texttt{\{shu.wu,wangliang\}@nlpr.ia.ac.cn} \\
 }

\begin{document}
\maketitle
\begin{abstract}
Hallucination has emerged as a significant barrier to the effective application of Large Language Models (LLMs).
In this work, we introduce a novel Attention-Guided SElf-Reflection (AGSER) approach for zero-shot hallucination detection in LLMs.
The AGSER method utilizes attention contributions to categorize the input query into attentive and non-attentive queries.
Each query is then processed separately through the LLMs, allowing us to compute consistency scores between the generated responses and the original answer.
The difference between the two consistency scores serves as a hallucination estimator.
In addition to its efficacy in detecting hallucinations, AGSER notably reduces computational overhead, requiring only three passes through the LLM and utilizing two sets of tokens.
We have conducted extensive experiments with four widely-used LLMs across three different hallucination benchmarks, demonstrating that our approach significantly outperforms existing methods in zero-shot hallucination detection.
\end{abstract}

\section{Introduction}
\label{sec:intro}

Recently, Large Language Models (LLMs) \cite{zhao2023survey} have demonstrated superior ability and achieved excellent results in various natural language processing tasks, such as summarization \cite{ravaut2024context}, machine translation \cite{zhang2023prompting}, autonomous agents \cite{wang2024survey}, information retrieval \cite{xuretrieval}, and knowledge graph reasoning \cite{sunthink}.
Despite the convenience offered by LLMs, they may produce overly confident answers that deviate from factual reality \cite{manakul2023selfcheckgpt,zhang2023sac3,he2024llm}.
This is usually called the \textit{Hallucination} phenomenon, which makes LLMs very untrustworthy \cite{zhang2023siren,li2024dawn,sun2025divide}.
This strongly limits the application of LLMs, especially in medical, financial, legal, and other scenarios.
Thus, it is urgent to investigate the accurate and efficient hallucination detection in LLMs, and teach LLMs to say ``I don't know'' when they are not sure about the answers.

The most common hallucination detection methods are based on answer consistency \cite{manakul2023selfcheckgpt,zhang2023sac3,cheninside}, in which the answers to the same query are sampled multiple times.
Though effective, such methods heavily increase computation cost through multiple LLM running.
They also rely on randomness, and when the LLM is extremely confident in the wrong answer, the same answer may be constantly generated during resampling \cite{zhang2023sac3}.
Moreover, none of the existing consistency-based approaches guides LLMs to rethink the answer generation process like humans do, which may help us to obtain a better consistency evaluation.
Recently, more hallucination detection approaches have been proposed from other perspectives, but they require tool usage \cite{cheng2024small}, or annotated hallucination datasets \cite{azaria2023internal,he2024llm,chuang2024lookback}.

Considering that attention contributions in LLMs reflect the key parts of the answer generation process and provide hints about hallucinations \cite{yuksekgonulattention,chen2025mixture}, we propose an Attention-Guided SElf-Reflection (\textbf{AGSER}) approach for zero-shot hallucination detection in LLMs, which refers to identifying hallucinations without requiring specific training on annotated samples from the target LLM.
Specifically, according to attention contributions of tokens, we split the input query for LLMs into attentive and non-attentive queries.
As the attentive query contains the major information for LLMs to generate the answer, if we input the attentive query into LLMs, the generated answer should be very similar to the original answer for a non-hallucination sample.
On the other hand, due to language differences between attentive and original queries, the randomness of generating the hallucination answer has been enlarged, and we have a greater chance of detecting hallucination based on the inconsistency of answers.
This is similar as when a human is doing reading comprehension, if asked to rethink about the answer, he or she will re-examine the attentive parts of the article, and may provide a new answer.
Meanwhile, for a non-hallucination sample, there is almost no important information in the non-attentive query, and thus when we input the non-attentive query into LLMs, the generated answer should be extremely random and totally different from the original answer.
In Sec. \ref{sec:analysis}, we provide some experimental observations to verify the above analysis.

Accordingly, in AGSER, we use attentive and non-attentive queries to guide LLMs to conduct self-reflection for hallucination detection.
Specifically, we separately feed attentive and non-attentive queries into LLMs, and respectively calculate the consistency scores between the generated answers and the original answer, which are denoted as attentive and non-attentive consistency scores.
Then, as smaller attentive consistency scores and larger non-attentive consistency scores indicate higher degrees of hallucination, we compute their difference as the hallucination estimator.
This enables us to detect hallucinations in a zero-shot manner.
Meanwhile, compared to conventional consistency-based approaches, AGSER reduces the computational overhead of resampling.
It only requires three times of LLM running, and two times of token usage.
We have conducted extensive experiments with four popular LLMs, and ASGER achieves state-of-the-art hallucination detection performances.

The main contributions of this work are summarized as follows:
\begin{itemize}
\item According to attention contributions of tokens in LLMs, we define attentive and non-attentive queries. For a hallucination sample, the generated answer of the attentive query has a larger chance to be different from the original answer, and the generated answer of the non-attentive query has a larger chance to be similar to the original answer.
\item We propose a novel AGSER approach for zero-shot hallucination detection. AGSER uses attentive and non-attentive queries for constructing an effective hallucination estimator. It can also reduce the computational overhead of answer resampling.
\item We have conducted extensive experiments with four popular LLMs, which demonstrate the effectiveness of our proposed AGSER approach in hallucination detection.
\end{itemize}

\section{Related Work}
\label{sec:related_work}

Hallucination has become the major obstacle in constructing trustworthy LLMs \cite{zhang2023siren}.
LLMs may generate overly confident non-factual contents.
This brings great demand for automatic hallucination detection in LLMs \cite{li2024dawn}, especially in a zero-shot manner.

\begin{figure*}[t]
  \centering
  \includegraphics[width=1\linewidth]{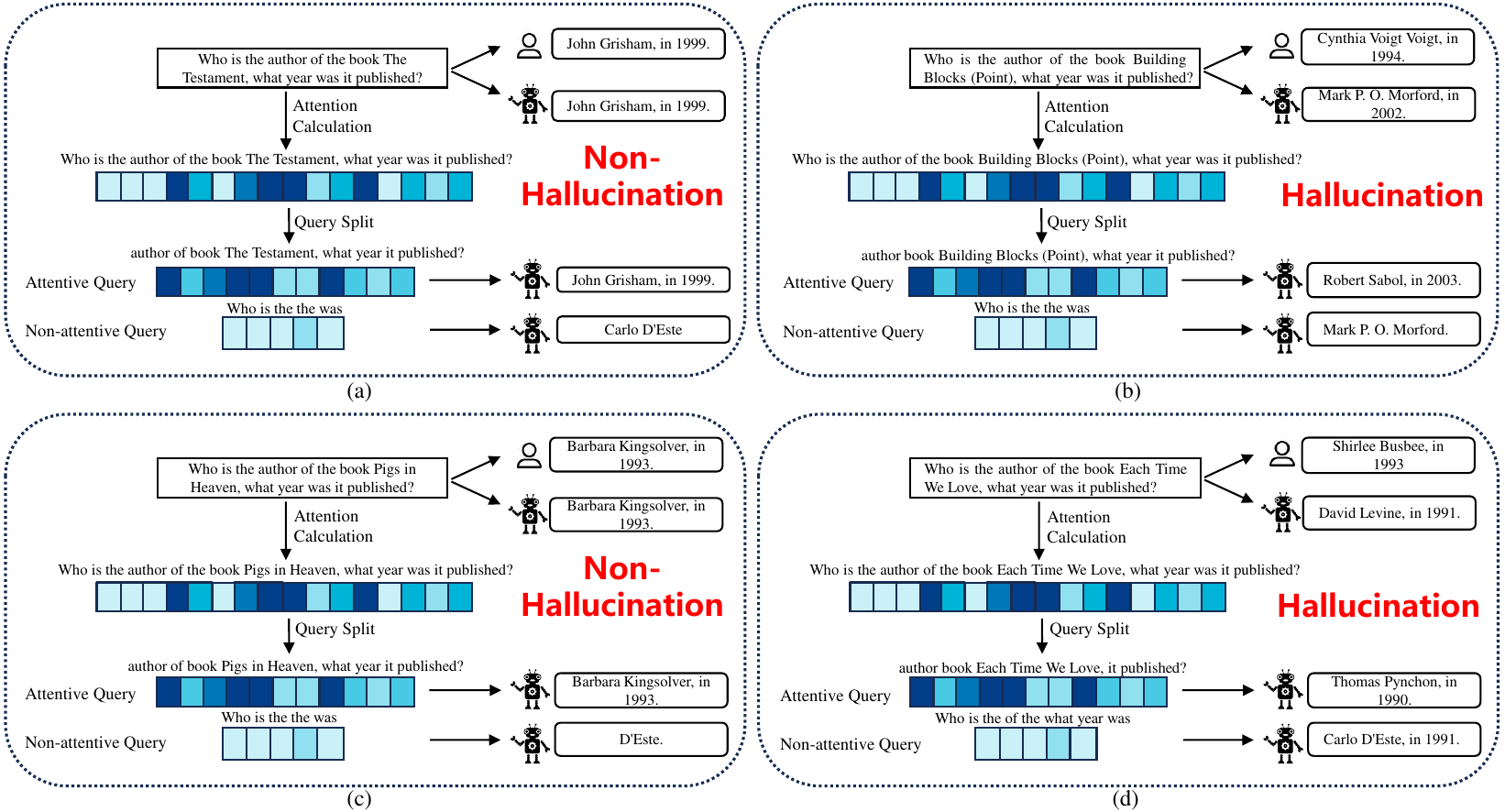}
  \caption{Some examples on feeding attentive and non-attentive queries into Llama2-7b. For non-hallucination samples, compared to the original answers, the answers of the attentive queries stay consistent, and those of the non-attentive queries otherwise. For hallucination samples, the answers of the attentive queries mostly change, and those of the non-attentive queries may remain unchanged.}
  \label{fig:examples}
\end{figure*}

The most common hallucination detection approach is based on the inconsistency of the generated contents.
SelfCheckGPT \cite{manakul2023selfcheckgpt} stochastically generates multiple responses besides the original answer, and detects the hallucination via verifying whether the responses support the original answer.
SAC$^3$ \cite{zhang2023sac3} detects hallucinations through consistency analysis across different LLMs or cross rephrased queries.
It also points out that generated answers to the same query may be consistent but non-factual.
LogicCheckGPT \cite{wu2024logical} asks LLMs with questions with logical relationships for hallucination detection.
INSIDE \cite{cheninside} attempts to calculate answer inconsistency in the sentence embedding space.
InterrogateLLM \cite{yehuda2024interrogatellm} detects hallucinations via asking the reverse question, and verify whether the original question can be generated.
Graph structure has also been extracted and applied for better estimation of answer consistency \cite{fang2025zero}.

Moreover, the inner states of LLMs can tell hallucinations to some extent \cite{azaria2023internal,zhong2025react}.
We can use hidden states \cite{he2024llm} or attention values \cite{chuang2024lookback} for training classifiers to detect hallucinations.
However, such approaches require training datasets, and may have trouble generalizing among different LLMs and different data \cite{orgad2024llms}.
Meanwhile, some works propose to call tools for constructing hallucination detectors \cite{cheng2024small,yin2023woodpecker}.
In addition, some works attempt to refine LLM parameters to enhance the factuality, via aligning with factuality analysis results \cite{zhang2024self}, truthful space editing \cite{zhang2024truthx}, over-trust penalty \cite{leng2024mitigating}, and confidence calibration \cite{liu2024enhancing}.
Contrastive decoding \cite{li2023contrastive,chuang2023dola,leng2024mitigating,cheng2025integrative,huo2025self}, which proposes to subtract output logits with less factuality, has also been used for improving the factuality.

There is research showing that, LLMs' attention to some constraint tokens (such as important entities) relates to the factuality of the generated responses \cite{yuksekgonulattention}.
Accordingly, attention contributions can reflect the answer generation process of LLMs, and guide LLMs to conduct self-reflection for accurate hallucination detection.

\section{Preliminary}
\label{sec:preliminary}

A query is denoted as a sequence of tokens $X = \left\{ {\mathop x\nolimits_1 ,\mathop x\nolimits_2 ,...,\mathop x\nolimits_M } \right\}$, in which $\mathop x\nolimits_i$ denotes the $i$-th token.
We denote a LLM as $f\left(  \bullet  \right)$, and the generated answer is $Y = f\left( X \right)$.
Specifically, the answer is a sequence of tokens $Y = \left\{ {\mathop y\nolimits_1 ,\mathop y\nolimits_2 ,...,\mathop y\nolimits_N } \right\}$, in which $\mathop y\nolimits_j$ denotes the $j$-th token.
Due to the hallucination phenomenon, $Y$ may be factual or non-factual.

The self-attention layers are the core components in LLMs \cite{vaswani2017attention}, and can reflect the key parts of the answer generation process of LLMs.
We assume that the LLM has $L$ self-attention layers and $H$ heads.
In the self-attention layers, there are two projection matrices $\mathop W\nolimits_Q^{l,h}$ and $\mathop W\nolimits_K^{l,h}$ for attention calculation, which denote query and key projections respectively, for layer $l$ and head $h$, and the dimensionality $d_h=d/H$.
The attention value matrix for layer $l$ and head $h$ can be calculated as
\begin{equation}
  \label{eq:attention_matrix}
  \small
  \mathop A\nolimits^{l,h}  = \sigma \left( {\frac{{\left( {\mathop X\nolimits^{l - 1} \mathop W\nolimits_Q^{l,h} } \right)\mathop {\left( {\mathop X\nolimits^{l - 1} \mathop W\nolimits_K^{l,h} } \right)}\nolimits^\top }}{{\sqrt {\mathop d\nolimits_h} }}} \right),
\end{equation}
where $\sigma$ denotes softmax function. And the attention contribution from token $j$ to token $i$ for layer $l$ through all heads can be calculated as
\begin{equation}
  \label{eq:attention_value}
  a_{i, j}^l = \sum_{h=1}^{H} A_{i, j}^{l, h}.
\end{equation}
Then, to obtain a score for measuring the contribution of the token $i$ during the answer generation process of the LLM, we use the attention contribution from token $i$ to the last token of the query as the token contribution score
\begin{equation}
  \label{eq:attention_score}
  \mathop s\nolimits_i^l  = \mathop a\nolimits_{M,i}^l.
\end{equation}

\section{Analysis}
\label{sec:analysis}

To verify that we can use attention to guide LLMs to conduct self-reflection and accurately detect hallucinations, we present the following analysis.
We adopt the attention at the middle layer, i.e., layer $L/2$, for the token contribution calculation. The contribution score at the middle layer for token $i$ is $\mathop s\nolimits_i^{mid}  = \mathop a\nolimits_{M,i}^{L/2}$, and the contribution scores for the entire input query are $\mathop S\nolimits^{mid}  = \left\{ {\mathop s\nolimits_1^{mid} ,...,\mathop s\nolimits_M^{mid} } \right\}$.
Then, we can split the input query $X = \left\{ {\mathop x\nolimits_1 ,\mathop x\nolimits_2 ,...,\mathop x\nolimits_M } \right\}$ into attentive and non-attentive queries
\begin{equation}
  \label{eq:attentive_query}
  \mathop X\nolimits^{att}  = \left\{ {\mathop x\nolimits_i |\mathop s\nolimits_i  \in top_k\left( {\mathop S\nolimits } \right)} \right\},
\end{equation}
\begin{equation}
  \label{eq:non_attentive_query}
  \mathop X\nolimits^{non\_att}  = \left\{ {\mathop x\nolimits_i |\mathop s\nolimits_i  \notin top_k\left( {\mathop S\nolimits } \right)} \right\},
\end{equation}
where $\mathop s\nolimits_i = \mathop s\nolimits_i^{mid}$, $\mathop S = \mathop S\nolimits^{mid}$, and $top_k\left(  \bullet  \right)$ means selecting tokens with $k$ highest contributions.
Here, we select the top $k=2/3$ tokens.
Then, we can obtain the corresponding responses of the LLM as $Y^{att} = f\left( X^{att} \right)$ and $Y^{non\_att} = f\left( X^{non\_att} \right)$.
To measure the consistency between the attention-guided generated answers $Y^{att}$, $Y^{non\_att}$ and the original answer $Y$, we adopt the Rouge-L \cite{lin2004rouge} similarity estimation \footnote{\url{https://github.com/google-research/google-research/tree/master/rouge}}, which provides an accurate evaluation for consistent answer pairs.
Specifically, we have attentive consistency score and non-attentive consistency score as follows
\begin{equation}
  \label{eq:attentive_sim}
  \mathop r\nolimits^{att}  = Rouge\left( {\mathop Y\nolimits^{att} ,Y} \right),
\end{equation}
\begin{equation}
  \label{eq:non_attentive_sim}
  \mathop r\nolimits^{non\_att}  = Rouge\left( {\mathop Y\nolimits^{non\_att} ,Y} \right).
\end{equation}

\begin{table}
  \centering
    \begin{tabular}{cccc}
    \toprule
    \multicolumn{4}{c}{Non-hallucination Samples} \\
    {[0.0,0.25)} & {[0.25,0.5)} & {[0.5,0.75)} & {[0.75,1.0]} \\
    0.025 & 0.167 & 0.218 & 0.590 \\
    \midrule
    \multicolumn{4}{c}{Hallucination Samples} \\
    {[0.0,0.25)} & {[0.25,0.5)} & {[0.5,0.75)} & {[0.75,1.0]} \\
    0.752 & 0.121 & 0.095 & 0.032 \\
    \bottomrule
    \end{tabular}%
  \caption{Distribution of attentive consistency scores $\mathop r\nolimits^{att}$ with Llama2-7b on the Books dataset.}
  \label{pilot-major}
\end{table}

\begin{table}
  \centering
    \begin{tabular}{cccc}
    \toprule
    \multicolumn{4}{c}{Non-hallucination Samples} \\
    {[0.0,0.25)} & {[0.25,0.5)} & {[0.5,0.75)} & {[0.75,1.0]} \\
    1.0 & 0.0 & 0.0 & 0.0 \\
    \midrule
    \multicolumn{4}{c}{Hallucination Samples} \\
    {[0.0,0.25)} & {[0.25,0.5)} & {[0.5,0.75)} & {[0.75,1.0]} \\
    0.845 & 0.121 & 0.031 & 0.003 \\
    \bottomrule
    \end{tabular}%
  \caption{Distribution of non-attentive consistency scores $\mathop r\nolimits^{non\_att}$ with Llama2-7b on the Books dataset.}
  \label{pilot-minor}
\end{table}

To analyze the relationship between hallucinations in LLMs and attentive/non-attentive consistency scores, we conduct some pilot study on the Books dataset \cite{yehuda2024interrogatellm}.
We present the results with the Llama2-7b model \cite{touvron2023llama}, which is a widely-used LLM.
In Fig. \ref{fig:examples}, we illustrate four examples on feeding attentive and non-attentive queries into Llama2-7b.
From the two non-hallucination samples we can observe that, the answers of the attentive queries stay consistent with the original answers, and the answers of the non-attentive queries are inconsistent with the original answers.
Meanwhile, as shown in the two hallucination samples, the answers of the attention queries mostly change, while the answers of the non-attentive queries may remain unchanged.
Furthermore, we show the distribution of attentive and non-attentive consistency scores in Tabs. \ref{pilot-major} and \ref{pilot-minor} respectively.
Obviously, the attentive consistency scores are much larger with non-hallucination samples than with hallucination samples.
Specifically, most attentive consistency scores of non-hallucination samples are in $[0.75, 1.0]$, while most attentive consistency scores of hallucination samples are in $[0.0,0.25)$.
Moreover, non-attentive consistency scores of non-hallucination samples are all in $[0.0,0.25)$,  while hallucination samples have the chance to have larger non-attentive consistency scores.
More results with other LLMs and on other datasets can be found in App. \ref{sec:pilot}.
We can conclude that, smaller attentive consistency scores and larger non-attentive consistency scores indicate greater probabilities of hallucinations.

\begin{algorithm}[t]
    \caption{The AGSER approach.}
    \label{alg:AGSER}
    \begin{algorithmic}[1]
        \REQUIRE A LLM $f\left(  \bullet  \right)$, and input query $X$.
        \ENSURE The degree of hallucination $r$.
        \STATE Feed the query $X$ into the LLM and obtain the answer $Y = f\left( X \right)$.
        \STATE Calculate the attention contributions in the LLM as in Eq. \ref{eq:attention_value}, and obtain the token contribution scores $\mathop S  = \left\{ {\mathop s\nolimits_1 ,...,\mathop s\nolimits_M } \right\}$.
        \STATE According to $\mathop S$, select the top $k$ tokens to construct the attentive query $\mathop X\nolimits^{att}$, and the rest to form the non-attentive query $\mathop X\nolimits^{non\_att}$ as in Eqs. \ref{eq:attentive_query} and \ref{eq:non_attentive_query}.
        \STATE Generate new answers $Y^{att} = f\left( X^{att} \right)$ and $Y^{non\_att} = f\left( X^{non\_att} \right)$.
        \STATE Calculate attentive and non-attentive consistency scores $\mathop r\nolimits^{att}$ and $\mathop r\nolimits^{non\_att}$ based on Rouge-L similarity estimation as in Eqs. \ref{eq:attentive_sim} and \ref{eq:non_attentive_sim}.
        \STATE Calculate the overall estimation of hallucination $r$ as in Eq. \ref{eq:final_score}.
        \RETURN $r$.
    \end{algorithmic}
\end{algorithm}

\section{Methodology}
\label{sec:method}

According to the above analysis and conclusion, in this section, we introduce the AGSER approach for zero-shot hallucination detection in LLMs.
The whole procedure is illustrated in Alg. \ref{alg:AGSER}.

In addition to adopting attention at the middle layer of a LLM for token contribution calculation as in Sec. \ref{sec:analysis}, we can define the following token contribution scores
\begin{itemize}
\item \textbf{The first layer value}: ${\rm{ }}s_i^{first} = {\rm{ }}a_{M,i}^1$.
\item \textbf{The middle layer value}: ${\rm{ }}s_i^{mid} = {\rm{ }}a_{M,i}^{L/2}$.
\item \textbf{The last layer value}: ${\rm{ }}s_i^{last} = {\rm{ }}a_{M,i}^L$.
\item \textbf{The maximum value of all layers}: \\$s_i^{max } = {\rm{ }}MAX\left( { {a_{M,i}^l} |0 < l \le L} \right)$.
\item \textbf{The mean value of all layers}: \\$s_i^{mean } = {\rm{ }}MEAN\left( { {a_{M,i}^l} |0 < l \le L} \right)$.
\end{itemize}
Then, we can replace the token contribution score $\mathop s\nolimits_i$ in Eqs. \ref{eq:attentive_query} and \ref{eq:non_attentive_query} with the above scores for calculating the corresponding attentive and non-attentive queries $\mathop X\nolimits^{att}$ and $\mathop X\nolimits^{non\_att}$.
And we can further obtain the attentive and non-attentive consistency scores $\mathop r\nolimits^{att}$ and $\mathop r\nolimits^{non\_att}$ for estimating the degrees of hallucinations in LLMs as in Eqs. \ref{eq:attentive_sim} and \ref{eq:non_attentive_sim}.

As smaller attentive consistency scores and larger non-attentive consistency scores indicate greater probabilities of hallucinations, we define the following score function as the final estimation of hallucinations in LLMs
\begin{equation}
  \label{eq:final_score}
  r = \lambda r^{att} - r^{non\_att}
\end{equation}
where $\lambda$ denotes a hyper-parameter for balancing the attentive and non-attentive consistency scores.
To be noted, lower scores indicate more severe hallucinations, and LLMs may generate non-factual contents.

\section{Experiments}
\label{sec:experiments}

In this section, we conduct extensive experiments to evaluate the effectiveness of AGSER in zero-shot hallucination detection in LLMs.

\subsection{Experimental Settings}
\label{sec:settings}

Following \cite{yehuda2024interrogatellm}, we conduct experiments on the \textbf{Books}, \textbf{Movies} and Global Country Information (\textbf{GCI}) datasets, which cover various domains.
For the evaluation of hallucination detection results, the detection
predictions are compared against the correctness of LLMs' answers.
The correctness is determined as in \cite{yehuda2024interrogatellm} for samples from different datasets.
More details of the datasets can be found in App. \ref{sec:data}.
Meanwhile, we use the Area Under Curve (\textbf{AUC}) as the evaluation metric.

We compare the proposed AGSER approach with SBERT~\cite{nils2019SBERT}, SelfCheckGPT~\cite{manakul2023selfcheckgpt}, INSIDE~\cite{cheninside} and InterrogateLLM~\cite{yehuda2024interrogatellm} in zero-shot hallucination detection.
Introduction of these baselines can be found in App. \ref{sec:baseline}.
Considering most inner state-based approaches require annotated dataset for training classifiers, we do not involve such approaches for comparison on zero-shot hallucination detection.

Moreover, we implement AGSER and other compared hallucination detection approaches with four popular and outstanding open-source LLMs: \textbf{Llama2-7b} \footnote{\url{https://huggingface.co/meta-llama/Llama-2-7b}}, \textbf{Llama2-13b} \footnote{\url{https://huggingface.co/meta-llama/Llama-2-13b}} \cite{touvron2023llama}, \textbf{Llama3-8b} \footnote{\url{https://huggingface.co/meta-llama/Meta-Llama-3-8B-Instruct}} \cite{dubey2024llama}, and \textbf{Qwen2.5-14b} \footnote{\url{https://huggingface.co/Qwen/Qwen2.5-14B-Instruct}} \cite{qwen2.5}.
More details of these LLMs can be found in App. \ref{sec:LLMs}.

For InterrogateLLM, we adopt the best version reported in the original paper, i.e., an ensemble of GPT-3 \cite{brown2020language}, Llama2-7b and Llama2-13b.
For SelfCheckGPT, INSIDE and InterrogateLLM, we perform resampling of answers for $5$ times to calculate the consistency scores.

\begin{table*}[t]
  \centering
    \begin{tabular}{c|ccc|ccc}
    \toprule
    \multirow{2}[2]{*}{Approaches} & \multicolumn{3}{c|}{Llama2-7b} & \multicolumn{3}{c}{Llama2-13b} \\
          & Books & Movies & GCI   & Books & Movies & GCI \\
    \midrule
    SBERT & 0.459 & 0.519 & 0.957 & 0.573 & 0.539 & 0.960 \\
    SelfCheckGPT & 0.783 & 0.811 & 0.790 & 0.751 & 0.794 & 0.885 \\
    INSIDE & 0.776 & 0.832 & 0.837 & 0.771 & 0.811 & 0.913 \\
    InterrogateLLM & 0.819 & 0.891 & 0.961 & 0.804 & 0.842 & 0.966 \\
    AGSER & \textbf{0.859} & \textbf{0.935} & \textbf{0.974} & \textbf{0.810} & \textbf{0.884} & \textbf{0.988} \\
    \midrule
    \multirow{2}[2]{*}{Approaches} & \multicolumn{3}{c|}{Llama3-8b} & \multicolumn{3}{c}{Qwen2.5-14b} \\
          & Books & Movies & GCI   & Books & Movies & GCI \\
    \midrule
    SBERT & 0.763 & 0.639 & 0.969 & 0.573 & 0.626 & 0.505 \\
    SelfCheckGPT & 0.825 & 0.802 & 0.721 & 0.711 & 0.763 & 0.607 \\
    INSIDE & 0.846 & 0.791 & 0.766 & 0.703 & 0.751 & 0.667 \\
    InterrogateLLM & 0.881 & 0.839 & \textbf{0.990} & 0.758 & 0.798 & 0.735 \\
    AGSER & \textbf{0.895} & \textbf{0.852} & 0.986 & \textbf{0.776} & \textbf{0.860} & \textbf{0.808} \\
    \bottomrule
    \end{tabular}%
  \caption{Performance comparison on zero-shot hallucination detection in LLMs.}
  \label{tab:performance}
\end{table*}

In our proposed AGSER approach, we set $k=2/3$ and $\lambda=1.0$.
And we adopt the mean value of all layers in a LLM, i.e., $s_i^{mean }$, for token contribution estimation.
We have not tuned the hyper-parameters for the optimal results on each dataset for each LLM, cause it is usually impractical to obtain sufficient high-quality hallucination and non-hallucination samples specific to each LLM as validation samples.
According to results in Sec. \ref{sec:params}, with the above selected hyper-parameters, we can not achieve the optimal results, but the overall satisfactory results.
Meanwhile, the prompts used in our experiments are illustrated in App. \ref{sec:prompts}.

\subsection{Performance Comparison}

The zero-shot hallucination detection results with four popular LLMs are illustrated in Tab. \ref{tab:performance}.
With different LLMs, similar comparison conclusions can be observed.
Not surprisingly, SBERT performs poorly, for it has no special design for measuring hallucinations in LLMs.
Detecting hallucinations in output space and embedding space respectively, SelfCheckGPT and INSIDE have similar detection results.
With detection AUC about $80\%$, they show their effectiveness in hallucination detection.
Meanwhile, via asking reverse questions, InterrogateLLM improves the detection results by large margins.
It allows the LLMs to rethink the generated answers from a new perspective, rather than only conducting multiple response resampling.
Moreover, obviously, compared to the above state-of-the-art approaches, our proposed AGSER approach achieves the best hallucination detection results.
With Llama2-7b, AGSER improves SelfCheckGPT, INSIDE and InterrogateLLM by $16.1\%$, $13.2\%$ and $3.6\%$ in average, respectively.
With Llama2-13b, AGSER improves SelfCheckGPT, INSIDE and InterrogateLLM by $10.4\%$, $7.5\%$ and $2.8\%$ in average, respectively.
With Llama3-8b, AGSER improves SelfCheckGPT, INSIDE and InterrogateLLM by $16.4\%$, $13.7\%$ and $0.9\%$ in average, respectively.
With Qwen2.5-14b, AGSER improves SelfCheckGPT, INSIDE and InterrogateLLM by $17.4\%$, $15.2\%$ and $6.7\%$ in average, respectively.
AGSER can significantly improve the detection performance with different LLMs across different datasets.
The only exception is evaluating with Llama3-8b on the GCI dataset, in which the detection AUC is nearly $1.0$.
These observations strongly demonstrate the superiority of using attention values to guide LLMs to conduct self-reflection for detecting hallucinations.

\begin{table*}[t]
  \centering
    \begin{tabular}{c|ccc|ccc}
    \toprule
    \multirow{2}[2]{*}{Approaches} & \multicolumn{3}{c|}{Llama2-7b} & \multicolumn{3}{c}{Llama2-13b} \\
          & Books & Movies & GCI   & Books & Movies & GCI \\
    \midrule
    AGSER & \textbf{0.859} & \textbf{0.935} & \textbf{0.974} & 0.810 & \textbf{0.884} & \textbf{0.988} \\
    AGSER w/ attentive queries & 0.848 & 0.926 & 0.970 & \textbf{0.814} & 0.875 & 0.984 \\
    AGSER w/ non-attentive queries & 0.572 & 0.581 & 0.545 & 0.508 & 0.649 & 0.631 \\
    \midrule
    \multirow{2}[2]{*}{Approaches} & \multicolumn{3}{c|}{Llama3-8b} & \multicolumn{3}{c}{Qwen2.5-14b} \\
          & Books & Movies & GCI   & Books & Movies & GCI \\
    \midrule
    AGSER & \textbf{0.895} & \textbf{0.852} & \textbf{0.986} & \textbf{0.776} & \textbf{0.860} & \textbf{0.808} \\
    AGSER w/ attentive queries & 0.887 & 0.846 & 0.984 & 0.765 & 0.846 & 0.800 \\
    AGSER w/ non-attentive queries & 0.553 & 0.556 & 0.511 & 0.581 & 0.625 & 0.589 \\
    \bottomrule
    \end{tabular}%
  \caption{Ablation study results regarding using only attentive or non-attentive queries for hallucination detection.}
  \label{tab:ablation-1}
\end{table*}

\begin{table*}[t]
  \centering
    \begin{tabular}{c|ccc|ccc}
    \toprule
    \multirow{2}[2]{*}{Approaches} & \multicolumn{3}{c|}{Llama2-7b} & \multicolumn{3}{c}{Llama2-13b} \\
          & Books & Movies & GCI   & Books & Movies & GCI \\
    \midrule
    AGSER w/ ${\rm{ }}s_i^{first}$ & 0.746 & 0.909 & 0.883 & 0.686 & 0.878 & 0.831 \\
    AGSER w/ ${\rm{ }}s_i^{mid}$ & 0.771 & 0.884 & \textbf{0.974} & 0.771 & \textbf{0.889} & 0.954 \\
    AGSER w/ ${\rm{ }}s_i^{last}$ & 0.792 & 0.849 & 0.962 & 0.741 & 0.815 & 0.973 \\
    AGSER w/ $s_i^{max }$ & 0.801 & 0.932 & 0.923 & 0.717 & 0.855 & 0.903 \\
    AGSER w/ $s_i^{mean } $ & \textbf{0.859} & \textbf{0.935} & \textbf{0.974} & \textbf{0.810} & 0.884 & \textbf{0.988} \\
    \midrule
    \multirow{2}[2]{*}{Approaches} & \multicolumn{3}{c|}{Llama3-8b} & \multicolumn{3}{c}{Qwen2.5-14b} \\
          & Books & Movies & GCI   & Books & Movies & GCI \\
    \midrule
    AGSER w/ ${\rm{ }}s_i^{first}$ & 0.727 & 0.790 & 0.862 & 0.669 & 0.779 & 0.765 \\
    AGSER w/ ${\rm{ }}s_i^{mid}$ & 0.848 & 0.843 & 0.941 & 0.676 & \textbf{0.882} & 0.761 \\
    AGSER w/ ${\rm{ }}s_i^{last}$ & 0.709 & 0.847 & 0.837 & 0.699 & 0.843 & 0.793 \\
    AGSER w/ $s_i^{max }$ & 0.753 & 0.815 & 0.979 & 0.756 & 0.836 & 0.762 \\
    AGSER w/ $s_i^{mean } $ & \textbf{0.895} & \textbf{0.852} & \textbf{0.986} & \textbf{0.776} & 0.860 & \textbf{0.808} \\
    \bottomrule
    \end{tabular}%
  \caption{Ablation study results regarding different token contribution scores.}
  \label{tab:ablation-2}
\end{table*}

\subsection{Ablation Study}

To investigate the effects of components and options in our proposed AGSER approach, we perform extensive ablation studies, and report the corresponding results.

Firstly, we investigate the effects of attentive and non-attentive queries, respectively.
Hallucination detection results of AGSER with only attentive queries or non-attentive queries are shown and compared to the results of AGSER in Tab. \ref{tab:ablation-1}.
Obviously, attentive query plays the major role in the effectiveness of AGSER.
And AGSER with only non-attentive queries achieves hallucination detection AUC of $0.575$ in average, which indicates non-attentive queries are also necessary for hallucination detection.
Specifically, without consideration of attentive queries, the detection AUC of AGSER decreases by $38.6\%$, $33.3\%$, $40.7\%$ and $26.6\%$ in average with the four LLMs respectively.
Meanwhile, without consideration of non-attentive queries, the detection AUC of AGSER decreases by $0.9\%$, $0.4\%$, $0.6\%$ and $1.4\%$ in average with the four LLMs respectively.
The above observations are reasonable, because only in a small portion of hallucination samples, the answers of non-attentive queries shall stay unchanged.
It is not an extremely strong indicator, but still a necessary one for reflecting the reasoning and answer generating process in LLMs.
In a word, both attentive and non-attentive queries are necessary and effective for detecting hallucinations in LLMs.

\begin{figure*}[t]
	\centering
	\subfigure[The Books dataset.]{
		\begin{minipage}[b]{0.315\textwidth}
			\includegraphics[width=1\textwidth]{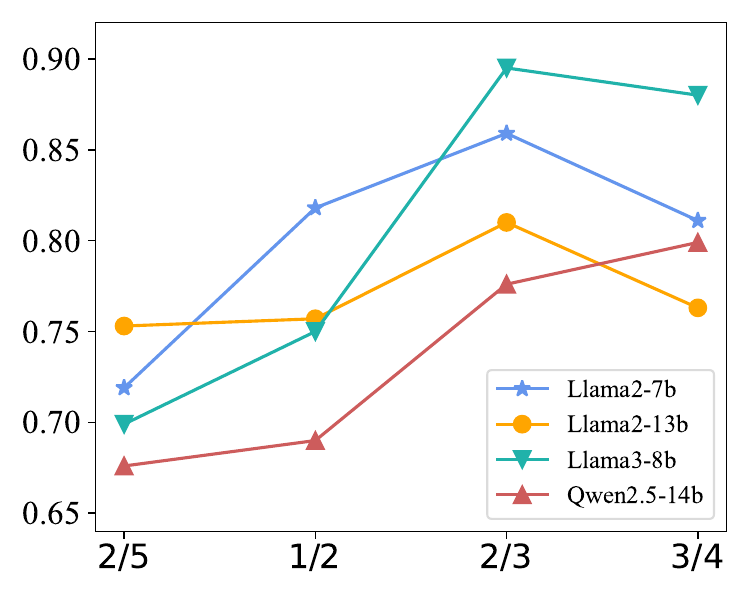}
		\end{minipage}
	\label{fig:params-1-1}
	}
	\subfigure[The Movies dataset.]{
		\begin{minipage}[b]{0.315\textwidth}
			\includegraphics[width=1\textwidth]{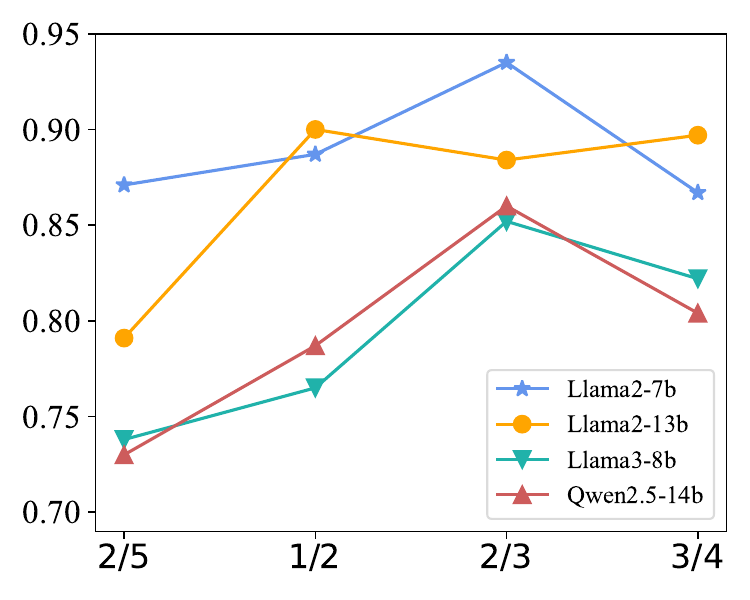}
		\end{minipage}
	\label{fig:params-1-2}
	}
	\subfigure[The GCI dataset.]{
		\begin{minipage}[b]{0.315\textwidth}
			\includegraphics[width=1\textwidth]{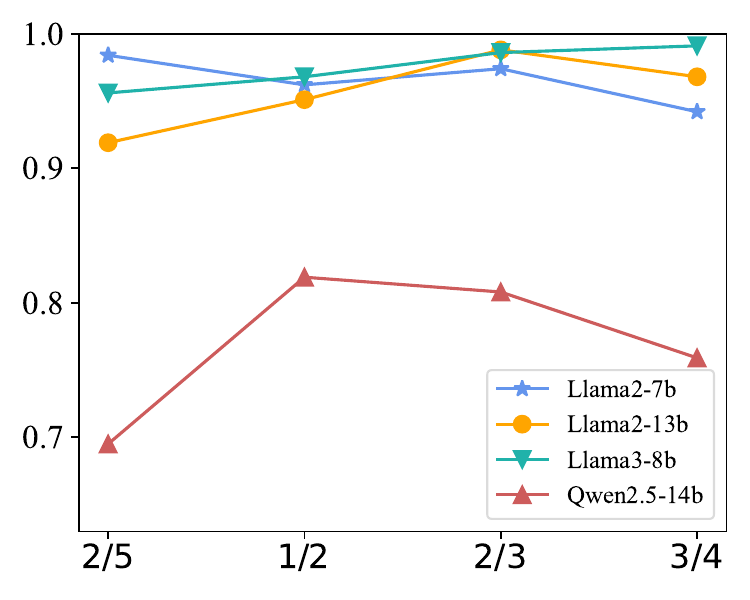}
		\end{minipage}
	\label{fig:params-1-3}
	}
	\caption{Hallucination detection results evaluated by AUC with varying $k$ values.}
	\label{fig:params-1}
\end{figure*}

\begin{figure*}[t]
	\centering
	\subfigure[The Books dataset.]{
		\begin{minipage}[b]{0.315\textwidth}
			\includegraphics[width=1\textwidth]{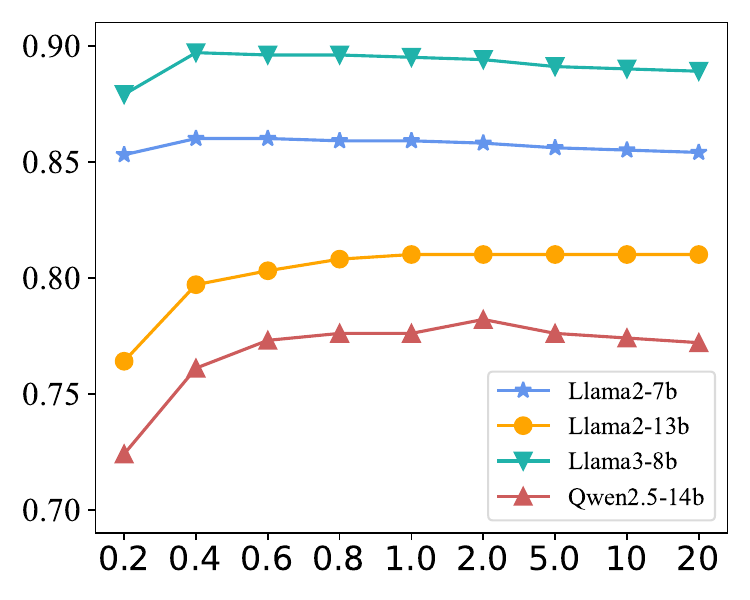}
		\end{minipage}
	\label{fig:params-2-1}
	}
	\subfigure[The Movies dataset.]{
		\begin{minipage}[b]{0.315\textwidth}
			\includegraphics[width=1\textwidth]{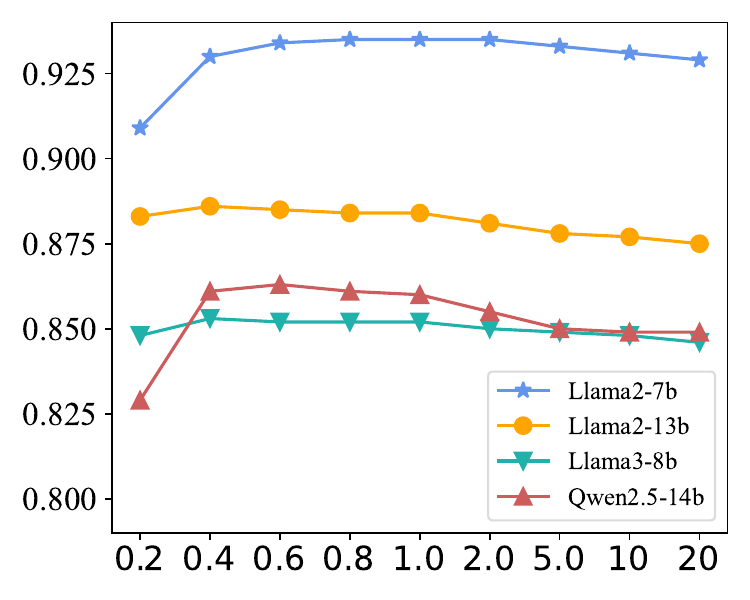}
		\end{minipage}
	\label{fig:params-2-2}
	}
	\subfigure[The GCI dataset.]{
		\begin{minipage}[b]{0.315\textwidth}
			\includegraphics[width=1\textwidth]{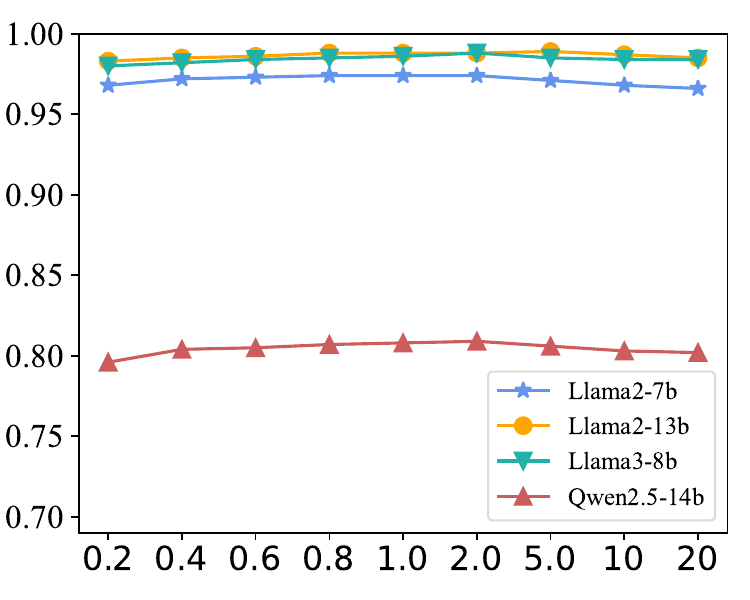}
		\end{minipage}
	\label{fig:params-2-3}
	}
	\caption{Hallucination detection results evaluated by AUC with varying $\lambda$ values.}
	\label{fig:params-2}
\end{figure*}

Secondly, we investigate the effects of different token contribution scores.
As introduced in Sec. \ref{sec:method}, there are five different token contribution scores: ${\rm{ }}s_i^{first}$, ${\rm{ }}s_i^{mid}$, ${\rm{ }}s_i^{last}$, $s_i^{max }$ and $s_i^{mean } $.
Accordingly, we report the hallucination detection results of AGSER with ${\rm{ }}s_i^{first}$, ${\rm{ }}s_i^{mid}$, ${\rm{ }}s_i^{last}$, $s_i^{max }$ and $s_i^{mean } $ respectively in Tab. \ref{tab:ablation-2}.
AGSER with ${\rm{ }}s_i^{first}$ achieves the lowest detection AUC of only $0.794$ in average.
Only considering the first layer attention contributions, we may lose some important states in the latter layers.
Considering the attention contributions in the last layer, which integrate some useful states in the formal layers, AGSER with ${\rm{ }}s_i^{last}$ achieves better detection AUC of $0.822$ in average.
Meanwhile, using the attention contributions in the middle layer, AGSER with ${\rm{ }}s_i^{mid}$ further improves the hallucination detection AUC to $0.849$ in average.
Moreover, with max pooling and mean pooling, we can capture the overall characteristics of all layers in LLMs more comprehensively, and thus achieve satisfactory hallucination detection results.
AGSER with $s_i^{max }$ and $s_i^{mean } $ achieves detection AUC of $0.836$ and $0.886$ in average, respectively.
Using the maximum values of all layers is obviously worse, indicating that max pooling may neglect some important information across different layers in LLMs.
Meanwhile, using the mean values of all layers is clearly better, and $s_i^{mean } $ is the best token contribution score according to our experimental results.

\subsection{Hyper-parameter Study}
\label{sec:params}

To investigate the impact of hyper-parameters in AGSER on the hallucination detection results, we conduct some hyper-parameter studies.
Firstly, we show the detection AUC with varying $k$ values in Fig. \ref{fig:params-1}.
The hyper-parameter $k$ controls the percentage of tokens selected for the attentive query.
In general, with larger $k$ values, which means retaining more sufficient major information in attentive queries, the results tend to be better.
But when $k=3/4$, in some cases, the detection results decrease slightly.
Secondly, we show the detection AUC with varying $\lambda$ values in Fig. \ref{fig:params-2}.
The hyper-parameter $\lambda$ controls the balance between attentive and non-attentive consistency scores.
In general, with different $\lambda$ values, the results are relatively stable.
Meanwhile, focusing too much on attentive or non-attentive consistency scores, AGSER will show some performance decline.

\subsection{Discussions}

According to the above observations, AGSER significantly outperforms state-of-the-art approaches on zero-shot hallucination detection in LLMs.
In addition, AGSER requires a lower computational overhead of resampling.
The compared methods, i.e., SelfCheckGPT, INSIDE and InterrogateLLM, perform $5$ times of LLM running.
In contrast, AGSER only requires $3$ times of LLM running (feeding original, attentive and non-attentive queries into LLMs), and $2$ times of token usage (attentive and non-attentive queries together have the same tokens as the original one).
In a word, AGSER has great advantages in both effectiveness and efficiency.
Furthermore, some running example results and bad cases of AGSER are presented in Apps. \ref{sec:examples} and \ref{sec:bad} respectively.

\section{Conclusion}


In summary, this work presents a systematic investigation of attention mechanisms in LLMs and proposes AGSER, a novel and computationally efficient approach for zero-shot hallucination detection. Through extensive experiments on three distinct factual knowledge recall tasks with four widely-used LLMs, AGSER demonstrates superior performance compared to existing hallucination detection methods.
Our findings make several key contributions to the field: (1) we provide new insights into how attention patterns correlate with hallucination behaviors in LLMs; (2) we establish AGSER as a robust and resource-efficient framework for hallucination detection. We believe that this work represents a significant step toward more reliable and trustworthy large language models.


\section*{Limitations}




While AGSER demonstrates promising results, we acknowledge several limitations of our approach. 

First, the method's reliance on attention allocation patterns during inference restricts its applicability to open-source LLMs, making it challenging to detect hallucinations in closed-source models accessed through APIs.

Furthermore, while AGSER achieves a remarkable $50\%$ or greater reduction in computational overhead compared to existing self-consistency methods, representing a significant breakthrough in efficiency, our approach still requires three inference passes with two token sets. The remaining computational requirements may still present challenges in specific scenarios, such as real-time applications or resource-constrained environments.

\section*{Ethical Considerations}


While our work aims to detect hallucinations, it is crucial to note that LLMs may still produce unreliable, biased, or factually incorrect information. Therefore, we emphasize that the outputs from our experimental results should be interpreted primarily as indicators of hallucination detection effectiveness rather than as reliable sources of factual information.

\section*{Acknowledgments}

This work is jointly sponsored by National Natural Science Foundation of China (62576339, 62141608, 62236010), Beijing Natural Science Foundation (L252033), and CAAI-Ant Group Research Fund.


\bibliography{acl_latex}

\begin{thebibliography}{39}
\providecommand{\natexlab}[1]{#1}

\bibitem[{Azaria and Mitchell(2023)}]{azaria2023internal}
Amos Azaria and Tom Mitchell. 2023.
\newblock The internal state of an llm knows when it's lying.
\newblock In \emph{Conference on Empirical Methods in Natural Language Processing}.

\bibitem[{Brown et~al.(2020)Brown, Mann, Ryder, Subbiah, Kaplan, Dhariwal, Neelakantan, Shyam, Sastry, Askell et~al.}]{brown2020language}
Tom Brown, Benjamin Mann, Nick Ryder, Melanie Subbiah, Jared~D Kaplan, Prafulla Dhariwal, Arvind Neelakantan, Pranav Shyam, Girish Sastry, Amanda Askell, et~al. 2020.
\newblock Language models are few-shot learners.
\newblock In \emph{Advances in Neural Information Processing Systems}.

\bibitem[{Chen et~al.(2024)Chen, Liu, Chen, Gu, Wu, Tao, Fu, and Ye}]{cheninside}
Chao Chen, Kai Liu, Ze~Chen, Yi~Gu, Yue Wu, Mingyuan Tao, Zhihang Fu, and Jieping Ye. 2024.
\newblock Inside: Llms' internal states retain the power of hallucination detection.
\newblock In \emph{International Conference on Learning Representations}.

\bibitem[{Chen et~al.(2025)Chen, Zhang, Liu, Wu, Zhang, and Tan}]{chen2025mixture}
Xinlong Chen, Yuanxing Zhang, Qiang Liu, Junfei Wu, Fuzheng Zhang, and Tieniu Tan. 2025.
\newblock Mixture of decoding: An attention-inspired adaptive decoding strategy to mitigate hallucinations in large vision-language models.
\newblock \emph{arXiv preprint arXiv:2505.17061}.

\bibitem[{Cheng et~al.(2024)Cheng, Li, Zhao, Zhang, Zhang, Zhang, Gai, and Wen}]{cheng2024small}
Xiaoxue Cheng, Junyi Li, Wayne~Xin Zhao, Hongzhi Zhang, Fuzheng Zhang, Di~Zhang, Kun Gai, and Ji-Rong Wen. 2024.
\newblock Small agent can also rock! empowering small language models as hallucination detector.
\newblock In \emph{Conference on Empirical Methods in Natural Language Processing}.

\bibitem[{Cheng et~al.(2025)Cheng, Liang, Gong, Xiao, Wang, Zhang, Hou, Xu, Liu, Li et~al.}]{cheng2025integrative}
Yi~Cheng, Xiao Liang, Yeyun Gong, Wen Xiao, Song Wang, Yuji Zhang, Wenjun Hou, Kaishuai Xu, Wenge Liu, Wenjie Li, et~al. 2025.
\newblock Integrative decoding: Improve factuality via implicit self-consistency.
\newblock In \emph{International Conference on Learning Representations}.

\bibitem[{Chuang et~al.(2024{\natexlab{a}})Chuang, Qiu, Hsieh, Krishna, Kim, and Glass}]{chuang2024lookback}
Yung-Sung Chuang, Linlu Qiu, Cheng-Yu Hsieh, Ranjay Krishna, Yoon Kim, and James Glass. 2024{\natexlab{a}}.
\newblock Lookback lens: Detecting and mitigating contextual hallucinations in large language models using only attention maps.
\newblock In \emph{Conference on Empirical Methods in Natural Language Processing}.

\bibitem[{Chuang et~al.(2024{\natexlab{b}})Chuang, Xie, Luo, Kim, Glass, and He}]{chuang2023dola}
Yung-Sung Chuang, Yujia Xie, Hongyin Luo, Yoon Kim, James~R Glass, and Pengcheng He. 2024{\natexlab{b}}.
\newblock Dola: Decoding by contrasting layers improves factuality in large language models.
\newblock In \emph{International Conference on Learning Representations}.

\bibitem[{Dubey et~al.(2024)Dubey, Jauhri, Pandey, Kadian, Al-Dahle, Letman, Mathur, Schelten, Yang, Fan et~al.}]{dubey2024llama}
Abhimanyu Dubey, Abhinav Jauhri, Abhinav Pandey, Abhishek Kadian, Ahmad Al-Dahle, Aiesha Letman, Akhil Mathur, Alan Schelten, Amy Yang, Angela Fan, et~al. 2024.
\newblock The llama 3 herd of models.
\newblock \emph{arXiv preprint arXiv:2407.21783}.

\bibitem[{Fang et~al.(2025)Fang, Huang, Tian, Fang, Pan, Fang, Wen, Pan, and Li}]{fang2025zero}
Xinyue Fang, Zhen Huang, Zhiliang Tian, Minghui Fang, Ziyi Pan, Quntian Fang, Zhihua Wen, Hengyue Pan, and Dongsheng Li. 2025.
\newblock Zero-resource hallucination detection for text generation via graph-based contextual knowledge triples modeling.
\newblock In \emph{AAAI Conference on Artificial Intelligence}.

\bibitem[{He et~al.(2024)He, Gong, Lin, Zhao, Chen et~al.}]{he2024llm}
Jinwen He, Yujia Gong, Zijin Lin, Yue Zhao, Kai Chen, et~al. 2024.
\newblock Llm factoscope: Uncovering llms’ factual discernment through measuring inner states.
\newblock In \emph{Findings of ACL}.

\bibitem[{Huo et~al.(2025)Huo, Xu, Zhang, Wang, Chen, and Zhao}]{huo2025self}
Fushuo Huo, Wenchao Xu, Zhong Zhang, Haozhao Wang, Zhicheng Chen, and Peilin Zhao. 2025.
\newblock Self-introspective decoding: Alleviating hallucinations for large vision-language models.
\newblock In \emph{International Conference on Learning Representations}.

\bibitem[{Leng et~al.(2024)Leng, Zhang, Chen, Li, Lu, Miao, and Bing}]{leng2024mitigating}
Sicong Leng, Hang Zhang, Guanzheng Chen, Xin Li, Shijian Lu, Chunyan Miao, and Lidong Bing. 2024.
\newblock Mitigating object hallucinations in large vision-language models through visual contrastive decoding.
\newblock In \emph{IEEE/CVF Conference on Computer Vision and Pattern Recognition}.

\bibitem[{Li et~al.(2024)Li, Chen, Ren, Cheng, Zhao, Nie, and Wen}]{li2024dawn}
Junyi Li, Jie Chen, Ruiyang Ren, Xiaoxue Cheng, Wayne~Xin Zhao, Jian-Yun Nie, and Ji-Rong Wen. 2024.
\newblock The dawn after the dark: An empirical study on factuality hallucination in large language models.
\newblock In \emph{Annual Meeting of the Association for Computational Linguistics}.

\bibitem[{Li et~al.(2023)Li, Holtzman, Fried, Liang, Eisner, Hashimoto, Zettlemoyer, and Lewis}]{li2023contrastive}
Xiang~Lisa Li, Ari Holtzman, Daniel Fried, Percy Liang, Jason Eisner, Tatsunori Hashimoto, Luke Zettlemoyer, and Mike Lewis. 2023.
\newblock Contrastive decoding: Open-ended text generation as optimization.
\newblock In \emph{Annual Meeting of the Association for Computational Linguistics}.

\bibitem[{Lin(2004)}]{lin2004rouge}
Chin-Yew Lin. 2004.
\newblock Rouge: A package for automatic evaluation of summaries.
\newblock In \emph{Text summarization branches out}, pages 74--81.

\bibitem[{Liu et~al.(2024)Liu, Bayat, and Wang}]{liu2024enhancing}
Xin Liu, Farima~Fatahi Bayat, and Lu~Wang. 2024.
\newblock Enhancing language model factuality via activation-based confidence calibration and guided decoding.
\newblock \emph{arXiv preprint arXiv:2406.13230}.

\bibitem[{Manakul et~al.(2023)Manakul, Liusie, and Gales}]{manakul2023selfcheckgpt}
Potsawee Manakul, Adian Liusie, and Mark Gales. 2023.
\newblock Selfcheckgpt: Zero-resource black-box hallucination detection for generative large language models.
\newblock In \emph{Conference on Empirical Methods in Natural Language Processing}.

\bibitem[{Orgad et~al.(2024)Orgad, Toker, Gekhman, Reichart, Szpektor, Kotek, and Belinkov}]{orgad2024llms}
Hadas Orgad, Michael Toker, Zorik Gekhman, Roi Reichart, Idan Szpektor, Hadas Kotek, and Yonatan Belinkov. 2024.
\newblock Llms know more than they show: On the intrinsic representation of llm hallucinations.
\newblock \emph{arXiv preprint arXiv:2410.02707}.

\bibitem[{Qwen(2024)}]{qwen2.5}
Team Qwen. 2024.
\newblock \href {https://qwenlm.github.io/blog/qwen2.5/} {Qwen2.5: A party of foundation models}.

\bibitem[{Ravaut et~al.(2024)Ravaut, Sun, Chen, and Joty}]{ravaut2024context}
Mathieu Ravaut, Aixin Sun, Nancy Chen, and Shafiq Joty. 2024.
\newblock On context utilization in summarization with large language models.
\newblock In \emph{Annual Meeting of the Association for Computational Linguistics}.

\bibitem[{Reimers and Gurevych(2019)}]{nils2019SBERT}
Nils Reimers and Iryna Gurevych. 2019.
\newblock Sentence-bert: Sentence embeddings using siamese bert-networks.
\newblock In \emph{Conference on Empirical Methods in Natural Language Processing}.

\bibitem[{Sun et~al.(2024)Sun, Xu, Tang, Wang, Lin, Gong, Ni, Shum, and Guo}]{sunthink}
Jiashuo Sun, Chengjin Xu, Lumingyuan Tang, Saizhuo Wang, Chen Lin, Yeyun Gong, Lionel Ni, Heung-Yeung Shum, and Jian Guo. 2024.
\newblock Think-on-graph: Deep and responsible reasoning of large language model on knowledge graph.
\newblock In \emph{International Conference on Learning Representations}.

\bibitem[{Sun et~al.(2025)Sun, Xie, Chen, Liu, Wu, Chen, Song, Wang, Wang, and Wang}]{sun2025divide}
Xin Sun, Jianan Xie, Zhongqi Chen, Qiang Liu, Shu Wu, Yuehe Chen, Bowen Song, Weiqiang Wang, Zilei Wang, and Liang Wang. 2025.
\newblock Divide-then-align: Honest alignment based on the knowledge boundary of rag.
\newblock In \emph{Annual Meeting of the Association for Computational Linguistics}.

\bibitem[{Touvron et~al.(2023)Touvron, Martin, Stone, Albert, Almahairi, Babaei, Bashlykov, Batra, Bhargava, Bhosale et~al.}]{touvron2023llama}
Hugo Touvron, Louis Martin, Kevin Stone, Peter Albert, Amjad Almahairi, Yasmine Babaei, Nikolay Bashlykov, Soumya Batra, Prajjwal Bhargava, Shruti Bhosale, et~al. 2023.
\newblock Llama 2: Open foundation and fine-tuned chat models.
\newblock \emph{arXiv preprint arXiv:2307.09288}.

\bibitem[{Vaswani et~al.(2017)Vaswani, Shazeer, Parmar, Uszkoreit, Jones, Gomez, Kaiser, and Polosukhin}]{vaswani2017attention}
Ashish Vaswani, Noam Shazeer, Niki Parmar, Jakob Uszkoreit, Llion Jones, Aidan~N Gomez, {\L}ukasz Kaiser, and Illia Polosukhin. 2017.
\newblock Attention is all you need.
\newblock In \emph{Advances in Neural Information Processing Systems}.

\bibitem[{Wang et~al.(2024)Wang, Ma, Feng, Zhang, Yang, Zhang, Chen, Tang, Chen, Lin et~al.}]{wang2024survey}
Lei Wang, Chen Ma, Xueyang Feng, Zeyu Zhang, Hao Yang, Jingsen Zhang, Zhiyuan Chen, Jiakai Tang, Xu~Chen, Yankai Lin, et~al. 2024.
\newblock A survey on large language model based autonomous agents.
\newblock \emph{Frontiers of Computer Science}, 18(6):186345.

\bibitem[{Wu et~al.(2024)Wu, Liu, Wang, Zhang, Wu, Wang, and Tan}]{wu2024logical}
Junfei Wu, Qiang Liu, Ding Wang, Jinghao Zhang, Shu Wu, Liang Wang, and Tieniu Tan. 2024.
\newblock Logical closed loop: Uncovering object hallucinations in large vision-language models.
\newblock \emph{arXiv preprint arXiv:2402.11622}.

\bibitem[{Xu et~al.(2024)Xu, Ping, Wu, McAfee, Zhu, Liu, Subramanian, Bakhturina, Shoeybi, and Catanzaro}]{xuretrieval}
Peng Xu, Wei Ping, Xianchao Wu, Lawrence McAfee, Chen Zhu, Zihan Liu, Sandeep Subramanian, Evelina Bakhturina, Mohammad Shoeybi, and Bryan Catanzaro. 2024.
\newblock Retrieval meets long context large language models.
\newblock In \emph{The Twelfth International Conference on Learning Representations}.

\bibitem[{Yehuda et~al.(2024)Yehuda, Malkiel, Barkan, Weill, Ronen, and Koenigstein}]{yehuda2024interrogatellm}
Yakir Yehuda, Itzik Malkiel, Oren Barkan, Jonathan Weill, Royi Ronen, and Noam Koenigstein. 2024.
\newblock Interrogatellm: Zero-resource hallucination detection in llm-generated answers.
\newblock In \emph{Annual Meeting of the Association for Computational Linguistics}.

\bibitem[{Yin et~al.(2023)Yin, Fu, Zhao, Xu, Wang, Sui, Shen, Li, Sun, and Chen}]{yin2023woodpecker}
Shukang Yin, Chaoyou Fu, Sirui Zhao, Tong Xu, Hao Wang, Dianbo Sui, Yunhang Shen, Ke~Li, Xing Sun, and Enhong Chen. 2023.
\newblock Woodpecker: Hallucination correction for multimodal large language models.
\newblock \emph{arXiv preprint arXiv:2310.16045}.

\bibitem[{Yuksekgonul et~al.(2024)Yuksekgonul, Chandrasekaran, Jones, Gunasekar, Naik, Palangi, Kamar, and Nushi}]{yuksekgonulattention}
Mert Yuksekgonul, Varun Chandrasekaran, Erik Jones, Suriya Gunasekar, Ranjita Naik, Hamid Palangi, Ece Kamar, and Besmira Nushi. 2024.
\newblock Attention satisfies: A constraint-satisfaction lens on factual errors of language models.
\newblock In \emph{International Conference on Learning Representations}.

\bibitem[{Zhang et~al.(2023{\natexlab{a}})Zhang, Haddow, and Birch}]{zhang2023prompting}
Biao Zhang, Barry Haddow, and Alexandra Birch. 2023{\natexlab{a}}.
\newblock Prompting large language model for machine translation: A case study.
\newblock In \emph{International Conference on Machine Learning}, pages 41092--41110.

\bibitem[{Zhang et~al.(2023{\natexlab{b}})Zhang, Li, Das, Malin, and Kumar}]{zhang2023sac3}
Jiaxin Zhang, Zhuohang Li, Kamalika Das, Bradley Malin, and Sricharan Kumar. 2023{\natexlab{b}}.
\newblock Sac3: Reliable hallucination detection in black-box language models via semantic-aware cross-check consistency.
\newblock In \emph{Findings of EMNLP}.

\bibitem[{Zhang et~al.(2024{\natexlab{a}})Zhang, Yu, and Feng}]{zhang2024truthx}
Shaolei Zhang, Tian Yu, and Yang Feng. 2024{\natexlab{a}}.
\newblock Truthx: Alleviating hallucinations by editing large language models in truthful space.
\newblock In \emph{Annual Meeting of the Association for Computational Linguistics}.

\bibitem[{Zhang et~al.(2024{\natexlab{b}})Zhang, Peng, Tian, Zhou, Jin, Song, Mi, and Meng}]{zhang2024self}
Xiaoying Zhang, Baolin Peng, Ye~Tian, Jingyan Zhou, Lifeng Jin, Linfeng Song, Haitao Mi, and Helen Meng. 2024{\natexlab{b}}.
\newblock Self-alignment for factuality: Mitigating hallucinations in llms via self-evaluation.
\newblock In \emph{Annual Meeting of the Association for Computational Linguistics}.

\bibitem[{Zhang et~al.(2023{\natexlab{c}})Zhang, Li, Cui, Cai, Liu, Fu, Huang, Zhao, Zhang, Chen et~al.}]{zhang2023siren}
Yue Zhang, Yafu Li, Leyang Cui, Deng Cai, Lemao Liu, Tingchen Fu, Xinting Huang, Enbo Zhao, Yu~Zhang, Yulong Chen, et~al. 2023{\natexlab{c}}.
\newblock Siren's song in the ai ocean: a survey on hallucination in large language models.
\newblock \emph{arXiv preprint arXiv:2309.01219}.

\bibitem[{Zhao et~al.(2023)Zhao, Zhou, Li, Tang, Wang, Hou, Min, Zhang, Zhang, Dong et~al.}]{zhao2023survey}
Wayne~Xin Zhao, Kun Zhou, Junyi Li, Tianyi Tang, Xiaolei Wang, Yupeng Hou, Yingqian Min, Beichen Zhang, Junjie Zhang, Zican Dong, et~al. 2023.
\newblock A survey of large language models.
\newblock \emph{arXiv preprint arXiv:2303.18223}.

\bibitem[{Zhong et~al.(2025)Zhong, Liu, Xu, Liu, Liu, Wu, Zhao, Wang, and Tan}]{zhong2025react}
Haitian Zhong, Yuhuan Liu, Ziyang Xu, Guofan Liu, Qiang Liu, Shu Wu, Zhe Zhao, Liang Wang, and Tieniu Tan. 2025.
\newblock React: Representation extraction and controllable tuning to overcome overfitting in llm knowledge editing.
\newblock In \emph{Conference on Empirical Methods in Natural Language Processing}.

\end{thebibliography}

\appendix

\section{Details of Datasets}
\label{sec:data}

We show the statistics of the Books, Movies and GCI datasets respectively in Tab. \ref{tab:data}.
In this work, as we aim to investigate the problem of zero-shot hallucination detection in LLMs, we use all the samples in the datasets for testing, and there are no training samples.

\section{More Pilot Study Results}
\label{sec:pilot}

Following the analysis in Sec. \ref{sec:analysis}, in this section, we present more pilot study results.
We provide more results with Llama2-7b, Llama2-13b, Llama3-8b and Qwen2.5-14b on the Books, Movies and GCI datasets.
The corresponding results are shown in Tabs. \ref{pilot-1}-\ref{pilot-22}.
We can draw the same conclusion as in Sec. \ref{sec:analysis}, i.e., smaller attentive consistency scores and larger non-attentive consistency scores indicate greater probabilities of hallucinations in LLMs.

\section{More Baseline Introduction}
\label{sec:baseline}

The compared zero-shot hallucination detection approaches are introduced as follows:
\begin{itemize}
    \item \textbf{SBERT}: Following \cite{yehuda2024interrogatellm}, we employ a pre-trained Sentence BERT model \cite{nils2019SBERT} as a baseline, which embeds both query and answer into vectors. Then, we calculate the cosine similarity between them as the hallucination prediction.
    \item \textbf{SelfCheckGPT}~\cite{manakul2023selfcheckgpt}: A detection approach that generates multiple responses and verifies whether they support the original answer.
    \item \textbf{INSIDE}~\cite{cheninside}: An approach that calculates eigenvalues of multiple answers in the sentence embedding space as the hallucination prediction estimator.
    \item \textbf{InterrogateLLM}~\cite{yehuda2024interrogatellm}: A state-of-the-art approach that detects hallucinations via feeding the reverse question into LLMs and verifies whether the original query could be generated.
\end{itemize}

\begin{table}
  \centering
    \begin{tabular}{c|ccc}
    \toprule
          & Books & Movies & GCI \\
    \midrule
    Number of Samples & 3000  & 3000  & 181 \\
    \bottomrule
    \end{tabular}%
  \caption{Statistics of the datasets.}
  \label{tab:data}
\end{table}

\section{More Detailed Settings}
\label{sec:LLMs}

The LLMs used in our experiments are introduced as follows:
\begin{itemize}
    \item \textbf{Llama 2-7B} is a variant of the Llama 2 family, and released in July 2023. It features 7 billion parameters, and is designed to perform a variety of natural language processing tasks.
    \item \textbf{Llama 2-13B} is also a variant of the Llama 2 family, and released in July 2023. It features 13 billion parameters.
    \item \textbf{Llama 3-8B} is a LLM from the Llama 3 series. It features 8 billion parameters, and is released in April 2024. It is one of the most advanced open-source LLMs.
    \item \textbf{Qwen 2.5-14B} is a LLM from the Qwen series. Released in September 2024, this model features 14 billion parameters. It is also one of the most advanced open-source LLMs, and shows great Chinese ability.
\end{itemize}

Moreover, all experiments are conducted on NVIDIA A100 GPUs with 80GB of memory.
We utilize a fixed random seed of 42, and the experimental results are reported within a single run.
Meanwhile, in our experiments, we employ the following versions of the libraries and models: SpaCy version 2.3.9, transformers version 4.30.2, and rouge version 1.0.1.

\section{Licensing}

The Books, Movies and GCI datasets are released for academic usage.
These datasets are designed for hallucination detection.
Thus, our use of these datasets is consistent with their intended use.

Moreover, Llama 2-7B and Llama 2-13B are released under the Meta Llama 2 Community License Agreement.
Llama 3-8B is released under the Meta Llama 3 Community License Agreement.
And Qwen 2.5-14B is released under the Apache-2.0 License.
They are all open for academic usage.

\section{Prompts}
\label{sec:prompts}

In this section, we detail the prompts for generating answers in LLMs.
The prompt template is shown in Fig. \ref{fig:prompts}.
And example prompts in the Books, Movies and GCI datasets are illustrated in Figs. \ref{fig:prompts-1}-\ref{fig:prompts-3} respectively.

\section{More Ablation Study Results}

In addition to the token contribution scores discussed in Sec. \ref{sec:method}, we investigate more layers in LLMs for token contribution calculation.
Results with different LLMs are shown in Tabs. \ref{ablation-x1}-\ref{ablation-x4}.
We can see that, AGSER w/ $s_i^{mean }$ can achieve the best overall performances.
And using values in some specific layers for calculating the token contribution scores can result in relatively high detection results in minor cases.

\section{Example Results}
\label{sec:examples}
In this section, we present some running example results of AGSER in Tabs. \ref{tab:example-1}-\ref{tab:example-8}.
We can observe that, for non-hallucination samples, compared to the original answers, the answers of the attentive queries stay consistent, and those of the non-attentive queries otherwise.
And for hallucination samples, the answers of the attentive queries mostly change, and those of the non-attentive queries may remain unchanged.
These observations enable our proposed AGSER approach to accurately detect hallucinations in LLMs.

\section{Bad Cases}
\label{sec:bad}

To investigate the shortage of AGSER and potential improvement, we demonstrate some bad cases:
\begin{itemize}
\item For the query ``Who is the author of the book Nights in Rodanthe, what year was it published?'', the LLM correctly responded with ``Nicholas Sparks, in 2002.'' However, the attentive query was incorrectly segmented as ``Nights in Rodanthe, what year?'', omitting the request for the author's name. Consequently, the LLM only answered ``In 2002,'' resulting in a final attentive consistency score of just $0.4$ for this non-hallucination sample.
\item Regarding the question ``Who is the author of the book Who Moved My Cheese?, what year was it published?'', the LLM erroneously answered ``Spencer Johnson, in 1996'' (the correct publication year being 1998). When the same question was posed as an attentive query, the response remained ``Spencer Johnson, in 1996,'' leading to an attentive consistency score of $0.99$ for this hallucination sample. This indicates that the LLM maintains incorrect memories about less commonly referenced information (such as book publication years).
\item For the query ``What actors played in the 1944 movie House of Frankenstein?'', the LLM initially provided the correct answer: ``The main cast included Boris Karloff, J. Carrol Naish and Lon Chaney Jr.'' However, the attentive query was mistakenly segmented as ``What actors played in the 1944 movie?'', omitting the movie title. This led the LLM to incorrectly respond with ``Peter Lorre,'' an actor active in the 1940s, resulting in an attentive consistency score of only $0.24$ for this non-hallucination sample.
\end{itemize}

Based on these bad cases, we can conclude that AGSER's erroneous judgments primarily stem from either incorrect segmentation of attentive queries (leading to omission of key information) or the LLM's inherent memory inaccuracies (especially for less commonly referenced information). These observations will help us further optimize our detection methods and develop more robust query segmentation strategies in future work.

\clearpage

\begin{table}[t]
  \centering
    \begin{tabular}{cccc}
    \toprule
    \multicolumn{4}{c}{Non-hallucination Samples} \\
    {[0.0,0.25)} & {[0.25,0.5)} & {[0.5,0.75)} & {[0.75,1.0]} \\
    0.092 & 0.130 & 0.212 & 0.566 \\
    \midrule
    \multicolumn{4}{c}{Hallucination Samples} \\
    {[0.0,0.25)} & {[0.25,0.5)} & {[0.5,0.75)} & {[0.75,1.0]} \\
    0.610 & 0.210 & 0.102 & 0.078    \\
    \bottomrule
    \end{tabular}%
  \caption{Distribution of attentive consistency scores $\mathop r\nolimits^{att}$ with Llama2-13b on the Books dataset.}
  \label{pilot-1}
\end{table}

\begin{table}[t]
  \centering
    \begin{tabular}{cccc}
    \toprule
    \multicolumn{4}{c}{Non-hallucination Samples} \\
    {[0.0,0.25)} & {[0.25,0.5)} & {[0.5,0.75)} & {[0.75,1.0]} \\
    0.989 & 0.011 & 0.0 & 0.0 \\
    \midrule
    \multicolumn{4}{c}{Hallucination Samples} \\
    {[0.0,0.25)} & {[0.25,0.5)} & {[0.5,0.75)} & {[0.75,1.0]} \\
    0.789 & 0.186 & 0.022 & 0.003 \\
    \bottomrule
    \end{tabular}%
  \caption{Distribution of non-attentive consistency scores $\mathop r\nolimits^{non\_att}$ with Llama2-13b on the Books dataset.}
  \label{pilot-2}
\end{table}

\begin{table}[t]
  \centering
    \begin{tabular}{cccc}
    \toprule
    \multicolumn{4}{c}{Non-hallucination Samples} \\
    {[0.0,0.25)} & {[0.25,0.5)} & {[0.5,0.75)} & {[0.75,1.0]} \\
    0.0 & 0.0 & 0.432 & 0.568 \\
    \midrule
    \multicolumn{4}{c}{Hallucination Samples} \\
    {[0.0,0.25)} & {[0.25,0.5)} & {[0.5,0.75)} & {[0.75,1.0]} \\
    0.822 & 0.108 & 0.007 & 0.063    \\
    \bottomrule
    \end{tabular}%
  \caption{Distribution of attentive consistency scores $\mathop r\nolimits^{att}$ with Llama3-8b on the Books dataset.}
  \label{pilot-3}
\end{table}

\begin{table}[t]
  \centering
    \begin{tabular}{cccc}
    \toprule
    \multicolumn{4}{c}{Non-hallucination Samples} \\
    {[0.0,0.25)} & {[0.25,0.5)} & {[0.5,0.75)} & {[0.75,1.0]} \\
    1.0 & 0.0 & 0.0 & 0.0 \\
    \midrule
    \multicolumn{4}{c}{Hallucination Samples} \\
    {[0.0,0.25)} & {[0.25,0.5)} & {[0.5,0.75)} & {[0.75,1.0]} \\
    0.986 & 0.012 & 0.001 & 0.001 \\
    \bottomrule
    \end{tabular}%
  \caption{Distribution of non-attentive consistency scores $\mathop r\nolimits^{non\_att}$ with Llama3-8b on the Books dataset.}
  \label{pilot-4}
\end{table}

\begin{table}[t]
  \centering
    \begin{tabular}{cccc}
    \toprule
    \multicolumn{4}{c}{Non-hallucination Samples} \\
    {[0.0,0.25)} & {[0.25,0.5)} & {[0.5,0.75)} & {[0.75,1.0]} \\
    0.127 & 0.181 & 0.262 & 0.430 \\
    \midrule
    \multicolumn{4}{c}{Hallucination Samples} \\
    {[0.0,0.25)} & {[0.25,0.5)} & {[0.5,0.75)} & {[0.75,1.0]} \\
    0.722 & 0.114 & 0.053 & 0.111    \\
    \bottomrule
    \end{tabular}%
  \caption{Distribution of attentive consistency scores $\mathop r\nolimits^{att}$ with Qwen2.5-14b on the Books dataset.}
  \label{pilot-5}
\end{table}

\begin{table}[t]
  \centering
    \begin{tabular}{cccc}
    \toprule
    \multicolumn{4}{c}{Non-hallucination Samples} \\
    {[0.0,0.25)} & {[0.25,0.5)} & {[0.5,0.75)} & {[0.75,1.0]} \\
    0.987 & 0.013 & 0.0 & 0.0 \\
    \midrule
    \multicolumn{4}{c}{Hallucination Samples} \\
    {[0.0,0.25)} & {[0.25,0.5)} & {[0.5,0.75)} & {[0.75,1.0]} \\
    0.907 & 0.070 & 0.015 & 0.008 \\
    \bottomrule
    \end{tabular}%
  \caption{Distribution of non-attentive consistency scores $\mathop r\nolimits^{non\_att}$ with Qwen2.5-14b on the Books dataset.}
  \label{pilot-6}
\end{table}

\begin{table}
  \centering
    \begin{tabular}{cccc}
    \toprule
    \multicolumn{4}{c}{Non-hallucination Samples} \\
    {[0.0,0.25)} & {[0.25,0.5)} & {[0.5,0.75)} & {[0.75,1.0]} \\
    0.051 & 0.165 & 0.189 & 0.595 \\
    \midrule
    \multicolumn{4}{c}{Hallucination Samples} \\
    {[0.0,0.25)} & {[0.25,0.5)} & {[0.5,0.75)} & {[0.75,1.0]} \\
    0.456 & 0.430 & 0.103 & 0.011 \\
    \bottomrule
    \end{tabular}%
  \caption{Distribution of attentive consistency scores $\mathop r\nolimits^{att}$ with Llama2-7b on the Movies dataset.}
  \label{pilot-7}
\end{table}

\begin{table}
  \centering
    \begin{tabular}{cccc}
    \toprule
    \multicolumn{4}{c}{Non-hallucination Samples} \\
    {[0.0,0.25)} & {[0.25,0.5)} & {[0.5,0.75)} & {[0.75,1.0]} \\
    1.0 & 0.0 & 0.0 & 0.0 \\
    \midrule
    \multicolumn{4}{c}{Hallucination Samples} \\
    {[0.0,0.25)} & {[0.25,0.5)} & {[0.5,0.75)} & {[0.75,1.0]} \\
    0.975 & 0.023 & 0.001 & 0.001 \\
    \bottomrule
    \end{tabular}%
  \caption{Distribution of non-attentive consistency scores $\mathop r\nolimits^{non\_att}$ with Llama2-7b on the Movies dataset.}
  \label{pilot-8}
\end{table}

\begin{table}[t]
  \centering
    \begin{tabular}{cccc}
    \toprule
    \multicolumn{4}{c}{Non-hallucination Samples} \\
    {[0.0,0.25)} & {[0.25,0.5)} & {[0.5,0.75)} & {[0.75,1.0]} \\
    0.026 & 0.117 & 0.320 & 0.537 \\
    \midrule
    \multicolumn{4}{c}{Hallucination Samples} \\
    {[0.0,0.25)} & {[0.25,0.5)} & {[0.5,0.75)} & {[0.75,1.0]} \\
    0.330 & 0.434 & 0.219 & 0.017    \\
    \bottomrule
    \end{tabular}%
  \caption{Distribution of attentive consistency scores $\mathop r\nolimits^{att}$ with Llama2-13b on the Movies dataset.}
  \label{pilot-9}
\end{table}

\begin{table}[t]
  \centering
    \begin{tabular}{cccc}
    \toprule
    \multicolumn{4}{c}{Non-hallucination Samples} \\
    {[0.0,0.25)} & {[0.25,0.5)} & {[0.5,0.75)} & {[0.75,1.0]} \\
    1.0 & 0.0 & 0.0 & 0.0 \\
    \midrule
    \multicolumn{4}{c}{Hallucination Samples} \\
    {[0.0,0.25)} & {[0.25,0.5)} & {[0.5,0.75)} & {[0.75,1.0]} \\
    0.864 & 0.128 & 0.007 & 0.001 \\
    \bottomrule
    \end{tabular}%
  \caption{Distribution of non-attentive consistency scores $\mathop r\nolimits^{non\_att}$ with Llama2-13b on the Movies dataset.}
  \label{pilot-10}
\end{table}

\begin{table}[t]
  \centering
    \begin{tabular}{cccc}
    \toprule
    \multicolumn{4}{c}{Non-hallucination Samples} \\
    {[0.0,0.25)} & {[0.25,0.5)} & {[0.5,0.75)} & {[0.75,1.0]} \\
    0.064 & 0.165 & 0.222 & 0.549 \\
    \midrule
    \multicolumn{4}{c}{Hallucination Samples} \\
    {[0.0,0.25)} & {[0.25,0.5)} & {[0.5,0.75)} & {[0.75,1.0]} \\
    0.442 & 0.357 & 0.192 & 0.009    \\
    \bottomrule
    \end{tabular}%
  \caption{Distribution of attentive consistency scores $\mathop r\nolimits^{att}$ with Llama3-8b on the Movies dataset.}
  \label{pilot-11}
\end{table}

\begin{table}[t]
  \centering
    \begin{tabular}{cccc}
    \toprule
    \multicolumn{4}{c}{Non-hallucination Samples} \\
    {[0.0,0.25)} & {[0.25,0.5)} & {[0.5,0.75)} & {[0.75,1.0]} \\
    1.0 & 0.0 & 0.0 & 0.0 \\
    \midrule
    \multicolumn{4}{c}{Hallucination Samples} \\
    {[0.0,0.25)} & {[0.25,0.5)} & {[0.5,0.75)} & {[0.75,1.0]} \\
    0.994 & 0.004 & 0.001 & 0.001 \\
    \bottomrule
    \end{tabular}%
  \caption{Distribution of non-attentive consistency scores $\mathop r\nolimits^{non\_att}$ with Llama3-8b on the Movies dataset.}
  \label{pilot-12}
\end{table}

\begin{table}[t]
  \centering
    \begin{tabular}{cccc}
    \toprule
    \multicolumn{4}{c}{Non-hallucination Samples} \\
    {[0.0,0.25)} & {[0.25,0.5)} & {[0.5,0.75)} & {[0.75,1.0]} \\
    0.121 & 0.152 & 0.303 & 0.424 \\
    \midrule
    \multicolumn{4}{c}{Hallucination Samples} \\
    {[0.0,0.25)} & {[0.25,0.5)} & {[0.5,0.75)} & {[0.75,1.0]} \\
    0.670 & 0.294 & 0.032 & 0.004    \\
    \bottomrule
    \end{tabular}%
  \caption{Distribution of attentive consistency scores $\mathop r\nolimits^{att}$ with Qwen2.5-14b on the Movies dataset.}
  \label{pilot-13}
\end{table}

\begin{table}[t]
  \centering
    \begin{tabular}{cccc}
    \toprule
    \multicolumn{4}{c}{Non-hallucination Samples} \\
    {[0.0,0.25)} & {[0.25,0.5)} & {[0.5,0.75)} & {[0.75,1.0]} \\
    1.0 & 0.0 & 0.0 & 0.0 \\
    \midrule
    \multicolumn{4}{c}{Hallucination Samples} \\
    {[0.0,0.25)} & {[0.25,0.5)} & {[0.5,0.75)} & {[0.75,1.0]} \\
    0.917 & 0.079 & 0.003 & 0.001 \\
    \bottomrule
    \end{tabular}%
  \caption{Distribution of non-attentive consistency scores $\mathop r\nolimits^{non\_att}$ with Qwen2.5-14b on the Movies dataset.}
  \label{pilot-14}
\end{table}

\begin{table}
  \centering
    \begin{tabular}{cccc}
    \toprule
    \multicolumn{4}{c}{Non-hallucination Samples} \\
    {[0.0,0.25)} & {[0.25,0.5)} & {[0.5,0.75)} & {[0.75,1.0]} \\
    0.0 & 0.0 & 0.013 & 0.987 \\
    \midrule
    \multicolumn{4}{c}{Hallucination Samples} \\
    {[0.0,0.25)} & {[0.25,0.5)} & {[0.5,0.75)} & {[0.75,1.0]} \\
    0.962 & 0.038 & 0.0 & 0.0 \\
    \bottomrule
    \end{tabular}%
  \caption{Distribution of attentive consistency scores $\mathop r\nolimits^{att}$ with Llama2-7b on the GCI dataset.}
  \label{pilot-15}
\end{table}

\begin{table}
  \centering
    \begin{tabular}{cccc}
    \toprule
    \multicolumn{4}{c}{Non-hallucination Samples} \\
    {[0.0,0.25)} & {[0.25,0.5)} & {[0.5,0.75)} & {[0.75,1.0]} \\
    1.0 & 0.0 & 0.0 & 0.0 \\
    \midrule
    \multicolumn{4}{c}{Hallucination Samples} \\
    {[0.0,0.25)} & {[0.25,0.5)} & {[0.5,0.75)} & {[0.75,1.0]} \\
    0.990 & 0.010 & 0.00 & 0.0 \\
    \bottomrule
    \end{tabular}%
  \caption{Distribution of non-attentive consistency scores $\mathop r\nolimits^{non\_att}$ with Llama2-7b on the GCI dataset.}
  \label{pilot-16}
\end{table}

\begin{table}[t]
  \centering
    \begin{tabular}{cccc}
    \toprule
    \multicolumn{4}{c}{Non-hallucination Samples} \\
    {[0.0,0.25)} & {[0.25,0.5)} & {[0.5,0.75)} & {[0.75,1.0]} \\
    0.0 & 0.0 & 0.080 & 0.920 \\
    \midrule
    \multicolumn{4}{c}{Hallucination Samples} \\
    {[0.0,0.25)} & {[0.25,0.5)} & {[0.5,0.75)} & {[0.75,1.0]} \\
    0.993 & 0.007 & 0.0 & 0.0    \\
    \bottomrule
    \end{tabular}%
  \caption{Distribution of attentive consistency scores $\mathop r\nolimits^{att}$ with Llama2-13b on the GCI dataset.}
  \label{pilot-17}
\end{table}

\begin{table}[t]
  \centering
    \begin{tabular}{cccc}
    \toprule
    \multicolumn{4}{c}{Non-hallucination Samples} \\
    {[0.0,0.25)} & {[0.25,0.5)} & {[0.5,0.75)} & {[0.75,1.0]} \\
    1.0 & 0.0 & 0.0 & 0.0 \\
    \midrule
    \multicolumn{4}{c}{Hallucination Samples} \\
    {[0.0,0.25)} & {[0.25,0.5)} & {[0.5,0.75)} & {[0.75,1.0]} \\
    0.840 & 0.120 & 0.020 & 0.020 \\
    \bottomrule
    \end{tabular}%
  \caption{Distribution of non-attentive consistency scores $\mathop r\nolimits^{non\_att}$ with Llama2-13b on the GCI dataset.}
  \label{pilot-18}
\end{table}

\begin{table}[t]
  \centering
    \begin{tabular}{cccc}
    \toprule
    \multicolumn{4}{c}{Non-hallucination Samples} \\
    {[0.0,0.25)} & {[0.25,0.5)} & {[0.5,0.75)} & {[0.75,1.0]} \\
    0.0 & 0.0 & 0.025 & 0.975 \\
    \midrule
    \multicolumn{4}{c}{Hallucination Samples} \\
    {[0.0,0.25)} & {[0.25,0.5)} & {[0.5,0.75)} & {[0.75,1.0]} \\
    0.986 & 0.014 & 0.0 & 0.0    \\
    \bottomrule
    \end{tabular}%
  \caption{Distribution of attentive consistency scores $\mathop r\nolimits^{att}$ with Llama3-8b on the GCI dataset.}
  \label{pilot-19}
\end{table}

\begin{table}[t]
  \centering
    \begin{tabular}{cccc}
    \toprule
    \multicolumn{4}{c}{Non-hallucination Samples} \\
    {[0.0,0.25)} & {[0.25,0.5)} & {[0.5,0.75)} & {[0.75,1.0]} \\
    1.0 & 0.0 & 0.0 & 0.0 \\
    \midrule
    \multicolumn{4}{c}{Hallucination Samples} \\
    {[0.0,0.25)} & {[0.25,0.5)} & {[0.5,0.75)} & {[0.75,1.0]} \\
    0.936 & 0.043 & 0.021 & 0.0 \\
    \bottomrule
    \end{tabular}%
  \caption{Distribution of non-attentive consistency scores $\mathop r\nolimits^{non\_att}$ with Llama3-8b on the GCI dataset.}
  \label{pilot-20}
\end{table}

\begin{table}[t]
  \centering
    \begin{tabular}{cccc}
    \toprule
    \multicolumn{4}{c}{Non-hallucination Samples} \\
    {[0.0,0.25)} & {[0.25,0.5)} & {[0.5,0.75)} & {[0.75,1.0]} \\
    0.011 & 0.011 & 0.024 & 0.954 \\
    \midrule
    \multicolumn{4}{c}{Hallucination Samples} \\
    {[0.0,0.25)} & {[0.25,0.5)} & {[0.5,0.75)} & {[0.75,1.0]} \\
    0.818 & 0.152 & 0.030 & 0.0    \\
    \bottomrule
    \end{tabular}%
  \caption{Distribution of attentive consistency scores $\mathop r\nolimits^{att}$ with Qwen2.5-14b on the GCI dataset.}
  \label{pilot-21}
\end{table}

\begin{table}[t]
  \centering
    \begin{tabular}{cccc}
    \toprule
    \multicolumn{4}{c}{Non-hallucination Samples} \\
    {[0.0,0.25)} & {[0.25,0.5)} & {[0.5,0.75)} & {[0.75,1.0]} \\
    0.994 & 0.006 & 0.0 & 0.0 \\
    \midrule
    \multicolumn{4}{c}{Hallucination Samples} \\
    {[0.0,0.25)} & {[0.25,0.5)} & {[0.5,0.75)} & {[0.75,1.0]} \\
    0.894 & 0.061 & 0.030 & 0.015 \\
    \bottomrule
    \end{tabular}%
  \caption{Distribution of non-attentive consistency scores $\mathop r\nolimits^{non\_att}$ with Qwen2.5-14b on the GCI dataset.}
  \label{pilot-22}
\end{table}

\begin{CJK*}{UTF8}{gbsn}
\begin{figure}[t]
\begin{tcolorbox}[colback=white!95!gray,colframe=gray!50!black,rounded corners,label={mid-zh-prompts-6}, title={Prompts}]
You are a helpful intelligent chatbot to answer questions.\\
Follow the format below, and please only predict the answer that corresponds to the last question.\\
Question: \{question\}\\
Answer:
\end{tcolorbox}
\caption{Prompts to answer the questions.}
\label{fig:prompts}
\end{figure}
\end{CJK*}

\begin{CJK*}{UTF8}{gbsn}
\begin{figure}[t]
\begin{tcolorbox}[colback=white!95!gray,colframe=gray!50!black,rounded corners,label={mid-zh-prompts-6}, title={Prompts}]
You are a helpful intelligent chatbot to answer questions.\\
Follow the format below, and please only predict the answer that corresponds to the last question.\\
Question: Who is the author of the book Classical Mythology, what year was it published?\\
Answer:
\end{tcolorbox}
\caption{Example prompts in the Books dataset.}
\label{fig:prompts-1}
\end{figure}
\end{CJK*}

\begin{CJK*}{UTF8}{gbsn}
\begin{figure}[t]
\begin{tcolorbox}[colback=white!95!gray,colframe=gray!50!black,rounded corners,label={mid-zh-prompts-6}, title={Prompts}]
You are a helpful intelligent chatbot to answer questions.\\
Follow the format below, and please only predict the answer that corresponds to the last question.\\
Question: What actors played in the 1995 movie Jumanji?\\
Answer:
\end{tcolorbox}
\caption{Example prompts in the Movies dataset.}
\label{fig:prompts-2}
\end{figure}
\end{CJK*}

\begin{CJK*}{UTF8}{gbsn}
\begin{figure}[t]
\begin{tcolorbox}[colback=white!95!gray,colframe=gray!50!black,rounded corners,label={mid-zh-prompts-6}, title={Prompts}]
You are a helpful intelligent chatbot to answer questions.\\
Follow the format below, and please only predict the answer that corresponds to the last question.\\
Question: What is the capital of France?\\
Answer:
\end{tcolorbox}
\caption{Example prompts in the GCI dataset.}
\label{fig:prompts-3}
\end{figure}
\end{CJK*}

\begin{table}[t]
  \centering
    \begin{tabular}{c|ccc}
    \toprule
    Layer & Books & Movies & GCI \\
    \midrule
    8     & 0.789 & 0.888 & 0.969 \\
    24    & 0.801 & 0.877 & 0.962 \\
    \bottomrule
    \end{tabular}%
  \caption{More ablation study results with Llama2-7b.}
  \label{ablation-x1}
\end{table}

\begin{table}[t]
  \centering
    \begin{tabular}{c|ccc}
    \toprule
    Layer & Books & Movies & GCI \\
    \midrule
    10     & 0.784 & 0.868 & 0.961 \\
    30    & 0.771 & 0.836 & 0.959 \\
    \bottomrule
    \end{tabular}%
  \caption{More ablation study results with Llama2-13b.}
  \label{ablation-x2}
\end{table}

\begin{table}[t]
  \centering
    \begin{tabular}{c|ccc}
    \toprule
    Layer & Books & Movies & GCI \\
    \midrule
    8     & 0.803 & 0.842 & 0.986 \\
    24    & 0.744 & 0.857 & 0.996 \\
    \bottomrule
    \end{tabular}%
  \caption{More ablation study results with Llama3-8b.}
  \label{ablation-x3}
\end{table}

\begin{table}[t]
  \centering
    \begin{tabular}{c|ccc}
    \toprule
    Layer & Books & Movies & GCI \\
    \midrule
    8     & 0.719 & 0.827 & 0.728 \\
    16    & 0.757 & 0.790 & 0.768 \\
    32    & 0.729 & 0.784 & 0.826 \\
    40    & 0.692 & 0.787 & 0.695 \\
    \bottomrule
    \end{tabular}%
  \caption{More ablation study results with Qwen2.5-14b.}
  \label{ablation-x4}
\end{table}

\begin{table*}[t]
  \centering
    \begin{tabular}{cl}
    \toprule
    Query & Who is the author of the book Dreamcatcher, what year was it published? \\
    Correct Answer & Stephen King, in 2001. \\
    Original Answer & Stephen King, in 2001. \\
    \midrule
    Attentive Query & author book Dreamcatcher, year it published? \\
    Answer & Stephen King, in 2001. \\
    \midrule
    Non-attentive Query & Who is the of the what was \\
    Answer & Carlo D'Este. \\
    \bottomrule
    \end{tabular}%
  \caption{AGSER's running example result 1.}
  \label{tab:example-1}
\end{table*}

\begin{table*}[t]
  \centering
    \begin{tabular}{cl}
    \toprule
    Query & Who is the author of the book Summer Sisters, what year was it published? \\
    Correct Answer & Judy Blume, in 1998. \\
    Original Answer & Judy Blume, in 1998. \\
    \midrule
    Attentive Query & book Summer Sisters, what it published? \\
    Answer & Judy Blume, in 1998. \\
    \midrule
    Non-attentive Query & Who is the author of the year was \\
    Answer & Mark P. O. Morford. \\
    \bottomrule
    \end{tabular}%
  \caption{AGSER's running example result 2.}
  \label{tab:example-2}
\end{table*}

\begin{table*}[t]
  \centering
    \begin{tabular}{cl}
    \toprule
    Query & Who is the author of the book Coraline, what year was it published? \\
    Correct Answer & Neil Gaiman, in 2002. \\
    Original Answer & Neil Gaiman, in 2002. \\
    \midrule
    Attentive Query & author book Coraline, year it published? \\
    Answer & Neil Gaiman, in 2002. \\
    \midrule
    Non-attentive Query & Who is the of the what was \\
    Answer & Carlo D'Este. \\
    \bottomrule
    \end{tabular}%
  \caption{AGSER's running example result 3.}
  \label{tab:example-3}
\end{table*}

\begin{table*}[t]
  \centering
    \begin{tabular}{cl}
    \toprule
    Query & Who is the author of the book At Home in Mitford, what year was it published? \\
    Correct Answer & Jan Karon, in 1996. \\
    Original Answer & Jan Karon, in 1996. \\
    \midrule
    Attentive Query & the author of the At Home in Mitford, was it published? \\
    Answer & Jan Karon, in 1996. \\
    \midrule
    Non-attentive Query & Who is book The what year \\
    Answer & The author of The Nightingale is Kristin Hannah, and it was published in 2015. \\
    \bottomrule
    \end{tabular}%
  \caption{AGSER's running example result 4.}
  \label{tab:example-4}
\end{table*}

\begin{table*}[t]
  \centering
    \begin{tabular}{cl}
    \toprule
    Query & Who is the author of the book Final Stand, what year was it published? \\
    Correct Answer & Helen Myers, in 2002. \\
    Original Answer & Mark P. O. Morford. \\
    \midrule
    Attentive Query & author of book Final Stand, what year it published? \\
    Answer & Michael Stephenson, in 2007. \\
    \midrule
    Non-attentive Query & Who is the the was  \\
    Answer & Mark P. O. Morford. \\
    \bottomrule
    \end{tabular}%
  \caption{AGSER's running example result 5.}
  \label{tab:example-5}
\end{table*}

\begin{table*}[t]
  \centering
    \begin{tabular}{cl}
    \toprule
    \multicolumn{1}{c}{\multirow{2}[0]{*}{Query}} & Who is the author of the book Secrets of St. John's Wort: A Lynn Sonberg Book,  \\
          & what year was it published? \\
    Correct Answer & Larry Katzenstein, in 1998. \\
    Original Answer & Lynn Sonberg, in 2003. \\
    \midrule
    Attentive Query & . John's Wort: A Lynn Sonberg Book,? \\
    Answer & 2001. \\
    \midrule
    Non-attentive Query & Who is the author of the book Secrets of St what year was it published  \\
    Answer & Mary's Hospital, in 2003. \\
    \bottomrule
    \end{tabular}%
  \caption{AGSER's running example result 6.}
  \label{tab:example-6}
\end{table*}

\begin{table*}[t]
  \centering
    \begin{tabular}{cl}
    \toprule
    Query & Who is the author of the book My Cat Spit McGee, what year was it published? \\
    Correct Answer & Willie Morris, in 1999. \\
    Original Answer & Mark P. O. Morford, in 2002. \\
    \midrule
    Attentive Query & author book My Cat Spit McGee, published? \\
    Answer & Iain Levison, in 2004. \\
    \midrule
    Non-attentive Query & Who is the of the what year was it  \\
    Answer & Mark P. O. Morford, in 2002. \\
    \bottomrule
    \end{tabular}%
  \caption{AGSER's running example result 7.}
  \label{tab:example-7}
\end{table*}

\begin{table*}[t]
  \centering
    \begin{tabular}{cl}
    \toprule
    \multicolumn{1}{c}{\multirow{2}[0]{*}{Query}} & Who is the author of the book Secrets of St. John's Wort: A Lynn Sonberg Book,  \\
          & what year was it published? \\
    Correct Answer & Marshall Kirk, in 1989. \\
    Original Answer & 1990 \\
    \midrule
    Attentive Query & book After Ball: Americaquerays in '90s, what year it published? \\
    Answer & 1999 \\
    \midrule
    Non-attentive Query & Who is the author of the the How Will Con Its Fear and Hatred of G the was  \\
    Answer & Thomas Pynchon, in 1990. \\
    \bottomrule
    \end{tabular}%
  \caption{AGSER's running example result 8.}
  \label{tab:example-8}
\end{table*}

\end{document}